\documentclass[11pt]{article}
\usepackage[letterpaper, margin=1in]{geometry}
\AtBeginDocument{%
  }

\usepackage{wrapfig}
\usepackage{microtype}
\usepackage{graphicx}
\usepackage{subfigure}
\usepackage{booktabs} 
\usepackage[utf8]{inputenc} 
\usepackage[T1]{fontenc}    
\usepackage{nicefrac} 
\usepackage{hyperref}
\hypersetup{
    colorlinks,
    linkcolor={red!50!black},
    citecolor={blue!50!black},
    urlcolor={blue!80!black}
}
\usepackage[sort, numbers]{natbib}

\usepackage{titlesec}
\usepackage{appendix}
\usepackage{xcolor}

\usepackage{amsmath, amssymb, amsthm, bbm,bm, mathtools,authblk}
\allowdisplaybreaks
\numberwithin{equation}{section}
\usepackage{enumerate}
\usepackage{float}
\mathtoolsset{showonlyrefs}
\usepackage{algorithm}
\usepackage{algorithmic}
\usepackage{tikz}
\usetikzlibrary{positioning}

\makeatletter
\newenvironment{subroutine}[1][htb]{%
    \renewcommand{\ALG@name}{Sub-Routine}
   \begin{algorithm}[#1]%
  }{\end{algorithm}}
\makeatother

\newtheorem{theorem}{Theorem}

\newtheorem{proposition}{Proposition}
\newtheorem{corollary}{Corollary}
\newtheorem{remark}{Remark}
\newtheorem{definition}{Definition}

\newcommand{\keywords}[1]{\textbf{\textit{Keywords ---}} #1}

\renewcommand{\P}{\mathbb{P}}
\newcommand{\E}{\mathbb{E}}
\newcommand{\R}{\mathbb{R}}
\newcommand{\N}{\mathbb{N}}
\newcommand{\F}{\mathcal{F}}

\DeclareMathOperator*{\argmin}{\arg\!\min}
\DeclareMathOperator*{\argmax}{\arg\!\max}

\newcommand{\ai}{{\texttt{LAI}}}

\newcommand{\ftm}{{\texttt{FTM}}}
\newcommand{\pol}{{\texttt{POL}}}
\newcommand{\scale}{{\texttt{SCaLE}}}
\newcommand{\hyscale}{{\texttt{HySCaLE}}}
\newcommand{\oal}{{\texttt{OAL}}}

\newcommand{\alg}{{\texttt{ALG}}}
\newcommand{\regret}{{\text{Regret}}}

\begin{document}

\title{SCaLE: Switching Cost aware Learning and Exploration}
\date{}

\author[1]{Neelkamal Bhuyan\thanks{Email: \href{mailto:nbhuyan3@gatech.edu}{nbhuyan3@gatech.edu}}\textsuperscript{,}}
\author[1]{Debankur Mukherjee}
\author[2]{Adam Wierman}

\affil[1]{Georgia Institute of Technology}
\affil[2]{California Institute of Technology}




\maketitle

\begin{abstract}
This work addresses the fundamental problem of unbounded metric movement costs in bandit online convex optimization, by considering high-dimensional dynamic quadratic hitting costs and $\ell_2$-norm switching costs in a noisy bandit feedback model. For a general class of stochastic environments, we provide the first algorithm \scale{} that provably achieves a \textit{distribution-agnostic sub-linear dynamic regret}, without the knowledge of hitting cost structure. En-route, we present a novel spectral regret analysis that separately quantifies eigenvalue-error driven regret and eigenbasis-perturbation driven regret. Extensive numerical experiments, against online-learning baselines, corroborate our claims, and highlight statistical consistency of our algorithm.
\end{abstract}

\section{Introduction}


Efficient management of large-scale infrastructure increasingly demands algorithms capable of learning system dynamics on-the-go while simultaneously maintaining optimal control. In many practical scenarios, pre-computed physical models are rendered obsolete by time-varying conditions or complex interactions, necessitating continuous online estimation. For instance, in data center thermal management and cooperative wind farm control, operators must estimate unknown coupling matrices, representing airflow turbulence or wake aerodynamics, without disrupting the stability of the active system. This challenge extends to smart grid topology identification, where impedance parameters fluctuate with load \cite{deka2012structure,cavraro2019real} robotics, where kinematic models are notoriously difficult to define analytically \cite{nguyen2011model}, and process control, where dynamics must be inferred safely in real-time \cite{hewing2020learning}. Section \ref{sec:applications} elaborates on a couple of such domains, where the learning agent faces a fundamental trade-off between probing the system to reduce model uncertainty, and minimizing the switching costs associated with rapid actuator fluctuations.

The interplay between system identification and control has been extensively studied to address this trade-off, historically within the framework of Adaptive Control. However, a significant limitation of the prevailing literature is its reliance on strict distributional assumptions. The majority of existing works, particularly those rooted in Linear Quadratic Gaussian (LQG) control, assume that the external driving force (or noise) is drawn from a sub-Gaussian distribution \cite{9483309,10590739}. While this assumption facilitates closed-form solutions via Riccati equations, it is frequently violated during practical deployment where disturbances may be heavy-tailed, structured, or adversarial. Consequently, algorithms derived under Gaussian priors often lack robustness when facing the erratic, non-stationary statistics of real-world demand or environmental load. This disconnect highlights a critical need for distribution-agnostic approaches that ensure performance stability without relying on specific noise generation models.

To bridge this gap, recent theoretical advances have connected the domains of Adaptive LQR control and Smoothed Online Convex Optimization (SOCO) \cite{GoelWierman19,BhuyanMukherjee24}. Specifically, \cite{BhuyanMukherjee24} establishes that for quadratic costs, one can derive a distribution-agnostic online optimal policy that mirrors the structure of classical LQR but remains valid for arbitrary martingale noise sequences, provided the system parameters are known. Building on this foundation, our work aims to extend stochastic system identification and control beyond the sub-Gaussian assumption. We address the problem where the governing cost matrix is unknown and must be learned from bandit feedback, effectively generalizing the robust SOCO framework to the adaptive setting. Specifically, our work seeks to answer the following question:

\begin{quote}
    \textit{How can we simultaneously (i) estimate an unknown cost function from noisy bandit feedback, (ii) use this estimate to make online decisions, (iii) account for switching-cost penalties when changing actions, and (iv) ensure minimal dynamic regret?}
\end{quote}

In addressing this question, we draw connections to the Bandit Convex Optimization (BCO) literature, which navigates dynamic costs with minimal feedback \cite{hazan2016optimal,sun2024tight,liu2025non}, and recent developments in Inverse Optimal Control (IOC), which utilize data-driven techniques to infer underlying cost structures \cite{lian2023off,asl2025data}. Section \ref{appendix_sec:lit_review} elaborates on the connection to literature in online algorithms, learning theory and optimal control.
 
\begin{wrapfigure}{r}{0.5\textwidth}
\vspace{-20pt}
    \centering
    \subfigure{%
        \includegraphics[width=\linewidth]{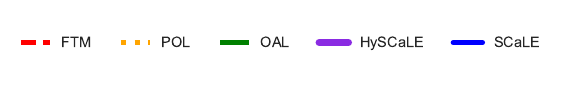}}\\
        \vspace{-25pt}
     \subfigure{%
        \includegraphics[width=0.48\linewidth]{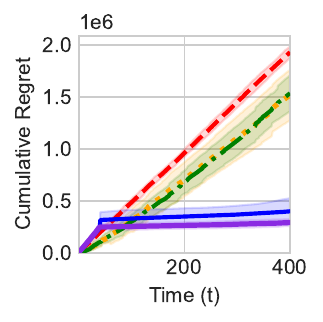}
        }
     \subfigure{%
    \includegraphics[width=0.48\linewidth]{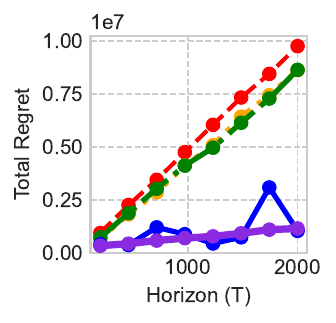}}
    \vspace{-10pt}
    \caption{\scale{} \& \hyscale{} performance (this work); \ftm{}: Smoothed online optimization benchmark \cite{zhang2021revisiting}; \pol{}/\oal{}: noisy/perfect online learning benchmark}
    \label{fig:intro_plots}
    \vspace{-15pt}
\end{wrapfigure}

The focus on switching costs is driven by the need for movement-efficient algorithms in large-scale sequential decision-making, including but not limited to smart grid cost management \cite{KimGiannakis17,WangHuang14}, adaptive control~\cite{Csaba11,Lale2020,Cohen2019,9483309,10590739},
data center management \cite{LinWierman12,albers2021algorithms}, electrical vehicle charging \cite{GanTopcu13}, video transmission \cite{ChenWierman2024} and power systems \cite{LuTuChau13,KimGiannakis17}. A core feature of many of these domains is the significant cost incurred when changing decisions, often exemplified by a continuous and unbounded space, owing to the fine-grained nature of decision parameters involved \cite{LinWierman12,zanini2013online,bhuyan2024optimal,ChenWierman2024}. 

However, control algorithms in the learning literature are often developed under assumptions that make the explicit management of movement costs a secondary concern. Their theoretical guarantees are typically developed for environments that either: (i) use static benchmarks \cite{bhaskara2021power,amir2022better} (ii) have a bounded action space diameter \cite{hazan2016optimal,zhao2023non}; (iii) are driven by low-drift, sub-gaussian stochastic processes \cite{10590739,9483309}; or (iv) satisfy certain controllability/stability conditions \cite{Lale2020,9483309,10590739}. The violation of any of these assumptions can lead to uncontrolled movement costs, a drawback of learning-based techniques illustrated in Figure \ref{fig:intro_plots}, and documented through case studies in Section \ref{subsec:case_studies}.

\vspace{.1in} \noindent\textbf{Challenges.}
In this paper, we focus on the class of quadratic hitting cost functions, motivated by the cost models employed by large-scale applications \cite{LinWierman12,KimGiannakis17,GoelWierman19,karapetyan2023online,BhuyanMukherjee24}.
At each round $t$, a player chooses an action $x_t \in \R^d$ and incurs a quadratic hitting cost $f_t(x) = \frac{1}{2}(x-v_t)^T A (x-v_t)$ and an $\ell_2$-norm switching cost $c(x_t, x_{t-1})=\frac{1}{2}\|x_t - x_{t-1}\|_2^2$. The unknown matrix $A \in \R^{d\times d}$ is positive semi-definite and, the minimizers of the hitting costs, $\{v_t\}_{t=1}^T$, which we refer to as \emph{location parameters}, are revealed in an online fashion, evolve as a stochastic process in $\R^d$ over finite time horizon. The esimate for this curvature matrix $A$ must be improved round-after-round in an online fashion, by observing only a single, noisy value of the incurred hitting cost, $f_t(x_t) + \eta_t$. With this noisy zeroth-order feedback, the agent faces two technical challenges:

\paragraph{(i) A Fundamental Trade-off.}
To effectively learn a $d\times d$ matrix $A$ from zeroth-order measurements, the learning agent must generate a sequence of sufficiently diverse query points $\{x_t\}_t$ \cite{moore2003persistence,simchowitz2020improper,lale2021model,muthirayan2022online,akbari2022achieving}. 
However, this requirement for exploration is in direct opposition to the objective of minimizing the switching cost $\|x_t-x_{t-1}\|_2^2$ which heavily penalizes large movements. This creates a novel tension not seen in standard bandit problems or online system identification: the very actions needed to gather information are explicitly discouraged by a core component of the objective function. Interestingly enough, this phenomenon shows up as a fundamental performance limit (Theorem \ref{thm:lower-bound}), and illustrated through a detailed example in Appendix \ref{sec:fundamental}.
\begin{wrapfigure}{r}{0.4\linewidth}
    \vspace{-15pt}
    \centering
     \subfigure[Curvature signal detection possible under noise]{%
        \includegraphics[width=\linewidth]{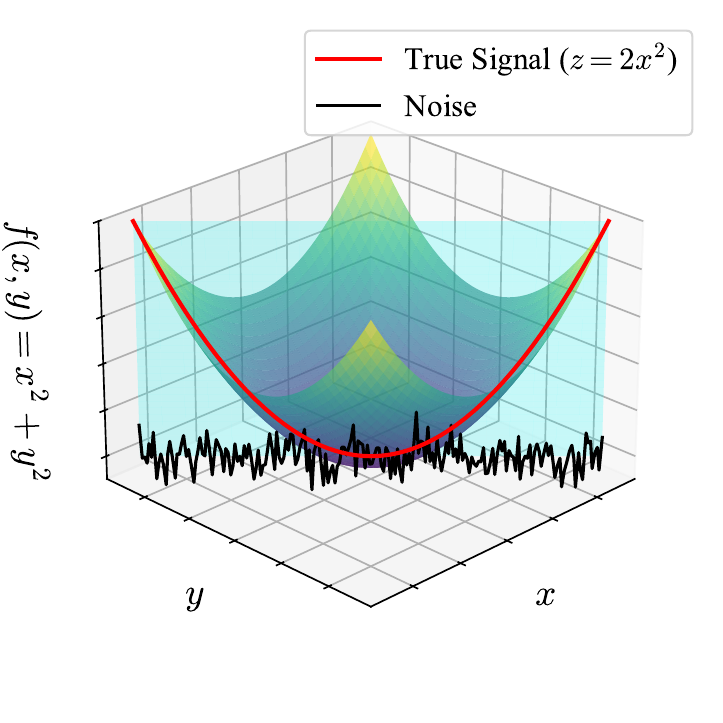}
        }\\
        \vspace{-9pt}
     \subfigure[Pure noise along directions of singularity]{%
    \includegraphics[width=\linewidth]{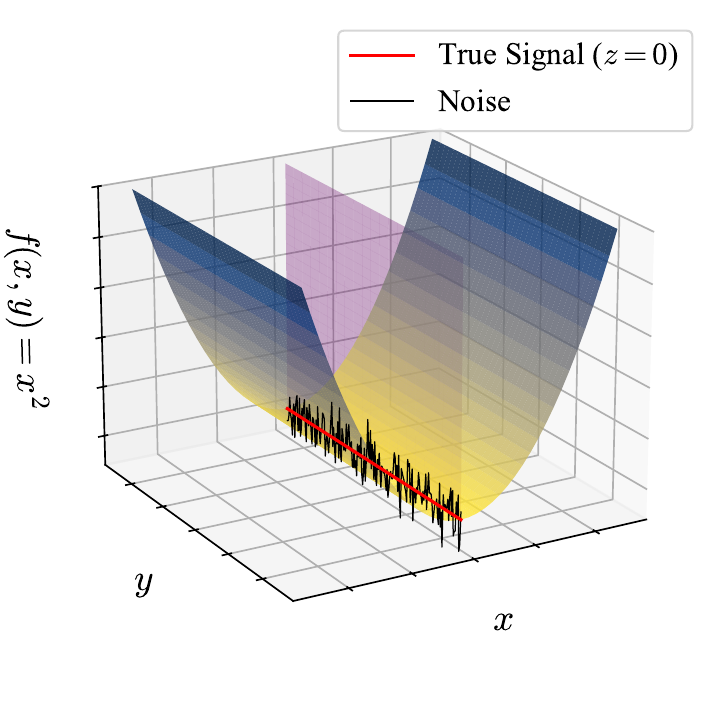}}
    \vspace{-10pt}
    \caption{Rank-deficiency hinders estimating curvature}
    \label{fig:3-d}
    \vspace{-25pt}
\end{wrapfigure}

\paragraph{(ii) Degeneracy in High Dimensions.}
This tension is exacerbated in high-dimensions when $f_t(x)$ exhibits regions of zero curvature (see Figure \ref{fig:3-d}). When $A$ is rank-deficient or singular, two critical failures occur: (i) the bandit feedback provides no useful signal for learning, effectively collapsing to pure noise, and (ii) switching costs overpower hitting costs. An agent might therefore incur a significant switching cost
only to receive a useless signal in return.

The above issue is a common roadblock in the literature on metric movement costs \cite{Borodin92,blum2000line,BansalGupta15,Antoniadis18,GoelLinWierman19,bhuyan2024optimal}. Prior work has established a strong $\Omega(\sqrt{d})$ competitive ratio lower bound in the adversarial setting, albeit through a pathological non-smooth construction \cite{Antoniadis18}. Practical settings, however, do not exhibit such extreme scenarios. In related domains, such as scheduling, stochastic modeling has been used to capture known data patterns, and algorithms can then be designed to perform well for these known models \cite{Koutsopoulos2011,Urgaonkar2011,Jennings1996,Gandhi2010,JosephVeciana12,11038902}.


This motivates our focus on the stochastic setting, where recent advances have demonstrated some promise. Notably, \cite{bhuyan2024optimal} circumvented this rank-deficiency barrier by leveraging the statistical properties of the minimizer's trajectory, albeit with full knowledge of the matrix $A.$ Building on this insight, we adopt the stochastic framework to address the combined challenges of estimation and movement in these degenerate, high-dimensional environments. 



\vspace{.1in}\noindent\textbf{Contributions.}
This paper makes four contributions:

\noindent(a)~\textit{Algorithmic: We design a novel algorithm, \scale{} (Algorithm \ref{alg:2}), that  achieves a distribution-agnostic sublinear dynamic regret} (Theorem \ref{thm:main_guarantees}) in an unbounded action space without any prior knowledge of the hitting cost structure.

\noindent(b)~\textit{Analytical:} \scale{}'s non-trivial performance is supported by rigorous performance guarantees:
\begin{enumerate}
\item An $\mathcal{O}(T^{2/3})$ dynamic regret for rank-deficient quadratic hitting costs, i.e., when rank$(A)<d$. This is the first high-dimensional guarantee in limited-feedback SOCO, that overcomes curvature-singularity (Theorem \ref{thm:exploit_regret_degen}).
\item Improved and tight $\Theta(T^{1/2})$ dynamic regret for the full-rank (rank$(A)=d$) setting (Theorem \ref{thm:exploit_regret_full}), matching the lower bound (Theorem \ref{thm:lower-bound}).
\end{enumerate}

\noindent(c)~\textit{A Novel Regret Decomposition:} These analytical results are enabled by key methodological advancements that might be of independent interest. We introduce:
\begin{enumerate}
\item A novel online regret decomposition (Theorem \ref{thm:regret_decomp}) that precisely separates the total regret into components characterized by: (i) statistical consistency, (ii) environment exploration and, (iii) exploitation.

\item A careful perturbation analysis (Proposition \ref{prop:spectral_split}) that uncovers a novel tight characterization of degenerate online quadratic optimization (Theorem \ref{thm:exploit_regret_degen} and Appendix \ref{proof:exploit_regret_degen}). This new technique separately quantifies the regret from estimation errors in the eigenbasis of the curvature matrix $A$ from errors in its eigenvalues.
\end{enumerate}

\noindent(d)~\textit{Empirical Insights:} Our experiments in Section \ref{sec:numerics} solidify the necessity of statistical consistency in dynamic estimation: Section \ref{subsec:case_studies} illustrates the failure of gradient-based techniques in heavy-tailed stochastic environments, but our proposed algorithm \scale{} still exhibits a sub-linear dynamic regret. For light-tailed environments, we propose a hybrid formulation \hyscale{} (Algorithm \ref{alg:2h}) by employing learning in the exploitation phase of \scale{}, resulting in (empirically) improved performance


\section{Model and Preliminaries}\label{sec:prelims}
Consider an online decision problem in an action space $\R^d$, $d \geq 1$, over a finite time horizon of $T$ rounds. In each round the player chooses an action $x_t$ and suffers a hitting cost of $f_t(x_t) = \frac{1}{2}(x_t-v_t)^T A (x_t-v_t)$, where $A$ is a positive semi-definite $d \times d$ matrix, and $v_t$ is a \emph{location parameter} that is revealed in an online fashion. Additionally, in each round, the player incurs a switching cost of $\frac{1}{2}\|x_t - x_{t-1}\|_2^2$ for transitioning between actions. We use $\|\cdot\|_*$ and $\|\cdot\|_F$ to denote the nuclear and Frobenius norm respectively, and $tr(\cdot)$ for the matrix trace operator.
\begin{definition}\label{defn:martingale_minimizer}
    The \textbf{stochastic environment} is defined by minimizer sequence $\{v_t\}_{t=1}^T$ as any martingale
    $$
        \E[v_t|\mathcal{F}_{t-1}] = v_{t-1} \text{ and } v_t-v_{t-1} \sim \mathrm{F}_t
    $$
    where $\{\mathcal{F}_s\}_{s=0}^T$ is the natural filtration generated by the minimizer sequence and the centered distributions $\{\mathrm{F}_s\}_{s=1}^T$ can be dynamic and light or heavy tailed.
\end{definition}

The stochastic setting we consider generalizes drift models in stochastic control literature, that considers additive white Gaussian noise driven dynamics~\cite{9483309,Csaba11,Lale2020,lale2021model,10590739}.


\begin{definition}\label{defn:bandit_feedback}
    For any $t$, we define \textbf{bandit feedback} or \textbf{rank-1 measurement} in the context of the player being aware of only the location $v_t$ prior to choosing $x_t$, after which it receives a noisy value of the hitting cost incurred, that is, $f_t(x_t)+ \eta_t$, where $\eta_t \in [-\bar{\eta},\bar{\eta}],$ can be random or adversarial.
\end{definition}
Under this information model, the player computes an online action $x^{\alg}_t$ at each round $t$ to solve the following optimization problem:
\begin{equation}\label{eqn:obj}
    \min_{(x_t)_{t=1}^T} \sum_{t=1}^T \frac{1}{2}(x_t-v_t)^T A (x_t-v_t) + \frac{1}{2}\|x_t-x_{t-1}\|_2^2
\end{equation}

Our choice of performance metric is as follows:
\begin{definition}\label{eqn:regret_defn}
    We define the \textbf{dynamic regret} of any online algorithm \alg{} is
    $$
        \regret^{\text{\alg{}}}_T := \E[\text{Cost}^{\text{\alg{}}}_{[1,T]}] - \E[\text{Cost}^*_{[1,T]}],
    $$
    where $\E[\text{Cost}^*[1,T]]$ is the total cost of the online optimal sequence of actions $\{x^*_t\}_{t=1}^T.$ 
\end{definition}

Note that the online optimal $\{x^*_t\}_{t=1}^T$ was characterized by \cite{BhuyanMukherjee24}, which solved \eqref{eqn:obj} through stochastic dynamic programming, albeit requiring exact knowledge of $A.$ Due to space constraints, detailed overview of related literature is deferred to Section \ref{appendix_sec:lit_review}.


    



\section{Algorithms and Regret Guarantees}
In this section, we propose our main algorithm \scale{} and establish its sub-linear dynamic regret in both full-rank and rank-deficient settings. 

\subsection{The \scale{} Algorithm}
\scale{} (Algorithm \ref{alg:2}) employs a carefully balanced \textbf{explore-then-exploit} strategy:

\begin{enumerate}
    \item \textbf{Exploration} (Lines 1-5): For rounds $\{1,\ldots,m\}$, \scale{} actively probes the cost function by injecting carefully scaled IID Gaussian noise $z_t$ around the revealed minimizer $v_t$. The goal here is to achieve \textbf{\textit{statistical consistency}} (see Remark \ref{remark: Cai-Zhang}) for which $m=c_1\cdot rd$. Although $c_1$ is a universal constant, we propose a line-search technique \eqref{eqn:c1_search}, for practical implementation.
    
    \item \textbf{Estimation and Exploitation} (Lines 6-14): After gathering sufficient data, the algorithm proceeds to the main estimation and control phase.
    
     (i) \noindent\textbf{Statistical Estimation} (Line 6): Solve for estimate, $\hat{A}^{\scale{}}$, by formulating the matrix recovery problem as a convex trace-norm minimization
        \begin{equation}\label{eqn: trace_min}
            \hat{A}^{\scale{}} = \argmin_{\substack{M \succeq 0\\ \|Y-\mathcal{A}(M)\|_1 \leq \bar{\eta}m}} \quad Tr(M) 
        \end{equation}
    (ii) \noindent\textbf{Planning} (Line 7): Using $\hat{A}^{\scale{}}$, \scale{} computes the \textit{entire} sequence of (near-)optimal matrices $\{\hat{C}_k\}_{k=1}^T$ via \texttt{RecurGen} (Subroutine \ref{subroutine:recursion}, provably optimal \cite{BhuyanMukherjee24} via dynamic programming).
    
    (iii) \noindent\textbf{Exploitation} (Lines 8-14): For the remaining $T-m$ rounds, \scale{} mimics the online-optimal \ai{} with (near-)optimal coefficients $\{\hat{C}_k\}_{k=m+1}^T$.
\end{enumerate}
\begin{algorithm}[]
    \caption{\scale{}}\label{alg:2}
    \flushleft \textbf{Input:}  noise cap $\bar{\eta}$, rank $r$, floor $\underline{\sigma_{r}^A}$, horizon $T$
    \flushleft \textbf{Initialize:} $m=c_1 r d$, $\hat{C}_{T+1} = I_{d\times d}$, $\gamma^2 = \begin{cases}
        \sqrt{\bar{\eta}}\max\left\{T^{2/3},\frac{1}{\underline{\sigma_{r}^A}}\right\} & r<d\\
        \sqrt{\bar{\eta}}\max\left\{T^{1/2},\frac{1}{\underline{\sigma_{r}^A}}\right\} & r=d
    \end{cases}$
    \begin{algorithmic}[1]
    \FOR{$t = 1,2,\ldots,m$}
    \STATE $z_t \sim \mathcal{N}(\mathbf{0},I_{d\times d})$
    \STATE $x_t \gets v_t + \gamma z_t$
    \STATE $y_t \gets f_t(x_t)+\eta_t$
    \ENDFOR
    \STATE $\hat{A}^{\scale{}} \gets \displaystyle\argmin_{\substack{M \succeq 0\\ \|Y-\mathcal{A}(M)\|_1 \leq \bar{\eta}m}} \quad Tr(M)$
    \STATE $\{\hat{C}_k\}_{k=1}^T \gets $ RecurGen$(\hat{A}^{\scale{}},1)$
    \FOR{$\tau=1,\ldots,m$}
        \STATE $x^{\ai{(\hat{A}^{\scale{}})}}_\tau \gets \hat{C}_\tau x^{\ai{(\hat{A}^{\scale{}})}}_{\tau-1} + (I-\hat{C}_\tau)v_\tau$
    \ENDFOR
    \STATE $x_{m+1} \gets \hat{C}_{m+1} x^{\ai{(\hat{A}^{\scale{}})}}_{m}+(I-\hat{C}_m)v_m$
    \FOR{$T = M+2,\ldots,T$}
        \STATE $x_t \gets \hat{C}_t x_{t-1}+(I-\hat{C}_t)v_t$
    \ENDFOR
    \end{algorithmic}
\end{algorithm}

\setcounter{algorithm}{0}
\begin{subroutine} [ht]
\caption{RecurGen}\label{subroutine:recursion}
\flushleft \textbf{Input}: Matrix $\hat{A}$, round $t$
\flushleft \textbf{Initialize:} $\hat{C}_{T+1} = I$
\begin{algorithmic}[1]
    \FOR{$\tau = T,\ldots,t$}
        \STATE $(\hat{C}_\tau)^{-1} = 2I + \hat{A}-\hat{C}_{\tau+1}$
    \ENDFOR\\
\RETURN $\{\hat{C}_k\}_{k=t}^T$
\end{algorithmic}
\end{subroutine}
\scale{} combines \eqref{eqn: trace_min}— a robust statistical tool originally for static matrix estimation \cite{7101247,CaiZhang15}, with distribution-agnostic online-optimal structure of \ai{} \cite{BhuyanMukherjee24}, using an explore-then-exploit strategy. The result is the following dynamic regret that holds even under heavy-tailed minimizer sequences $\{v_t\}_t$.
\begin{theorem}\label{thm:main_guarantees}
Dynamic regret of \scale{$(\bar{\eta},r,\underline{\sigma^A_{r}})$}, for $A$ with rank $r$, is bounded as:
$$
    \regret_T^{\text{\scale{}}} \leq
    \begin{cases}
             \begin{split}
                 \alpha_1(\sqrt{\bar{\eta}},\underline{\sigma^A_{r}}&,\Sigma,d)\boldsymbol{(T-c_1rd)^{2/3}} + \alpha_2(\sqrt{\bar{\eta}},\underline{\sigma^A_{r}},\Sigma,d)
             \end{split} & r<d\\
        \begin{split}
            \beta_1(\sqrt{\bar{\eta}},\underline{\sigma^A_{d}}&,\Sigma,d)\boldsymbol{(T-c_1d^2)^{1/2}} + \beta_2(\sqrt{\bar{\eta}},\underline{\sigma^A_{d}},\Sigma,d)
        \end{split} & r=d.
    \end{cases}
$$
with high probability ($1-\exp(-C_0 m)$), where $C_0,c_1$ are universal constants from matrix estimation theory \cite{7101247,CaiZhang15}, and $\Sigma$ is the upper limit on the covariance of the disturbance process $\{v_t-v_{t-1}\}_t$, that is,  $\Sigma_{v_t-v_{t-1}} \preceq \Sigma$. The terms $\alpha_1,\alpha_2,\beta_1,\beta_2 >0,$ with exact dependence in Appendix \ref{proof:equal-contribution}.
\end{theorem}


\begin{remark}\label{remark: Cai-Zhang}
     Lower bound \cite[Theorem 2.4]{CaiZhang15} on ``$m$'' establishes that $\Theta(r\cdot d)$ rank-$1$ measurements are necessary to ensure the ``Restricted Uniform Boundedness'' (RUB) property of the estimator $\hat{A}$, required for stable recovery under noise. In its absence, any estimator $\hat{A}$ is \textbf{\textit{statistically inconsistent}}, that is, 
     $$
        \inf_{\hat{A}} \sup_{\substack{A \in \R^{d\times d}\\rank(A)=r}} \E\|\hat{A}-A\|_F^2 = \infty.
     $$ 
     Section \ref{subsec:case_studies} illustrates the downstream effect of this inconsistency on the dynamic regret of online gradient descent based algorithms.
\end{remark}
In particular, there is no dependence on distribution other than the co-variance upper bound $tr(\Sigma).$ This tail-agnostic guarantee is the key feature that sets the above result apart from previous attempts in LQG control \cite{9483309,10590739}.

The distinct regret guarantees stem from two primary sources of spectral error—perturbation to the eigenvalues and, that to the eigenbasis of the estimated curvature matrix. We show in Section \ref{sec:proof_sketch} how the different scaling of these two error components, as a function of estimation error ($\epsilon$) and the horizon ($T$), is precisely what leads to the $\mathcal{O}(T^{2/3})$ rate in one case and the $\mathcal{O}(T^{1/2})$ rate in the other. In particular, Theorem \ref{thm:exploit_regret_degen} establishes a novel error rate that is an order of magnitude lower ($\epsilon^2$ instead of $\epsilon$) than what standard regret analysis from \cite{BhuyanMukherjee24} would find.

Furthermore, the $\mathcal{O}(T^{1/2})$ rate in the full-rank case confirms the optimality of our approach, as it matches the information-theoretic lower bound for this setting.

\subsection{Regret Lower Bound}
To establish the fundamental limits of the problem, we now complement our algorithmic upper bounds with a regret lower bound for a broad class of online algorithms—those that make decisions by interpolating between their previous action and the current minimizer through matrix $C_t \preceq I$:
$$
    x_t = C_t x_{t-1} + (I-C_t)v_t
$$
This structure is general, encompassing all state-of-the-art SOCO approaches for quadratic costs. The following theorem formalizes the inherent exploration cost that any such algorithm must pay when operating under the constraints of bandit feedback.
\begin{theorem}\label{thm:lower-bound}
    If an ``interpolation'' algorithm \alg{} in the above class performs rank-$1$ measurements to learn the optimal matrices $\{C^*_t\}_t$ to within an error of $\epsilon$ (i.e., $\frac{\epsilon}{2}\leq \sigma^{C_t-C^*_t}_{\min} \leq  \sigma^{C_t-C^*_t}_{\max}\leq \epsilon$), then there exists an ``$A$'' such that \alg{} suffers a regret of at least
     \begin{align}
         \regret_T^{\alg} = \frac{\alpha_1}{\epsilon}+\alpha_2\epsilon T +\alpha_3 = \Omega(\sqrt{T}),
     \end{align}
    where $\alpha_1,\alpha_2 \in \R^+$ and $\alpha_3\in \R$ are independent of $T.$
\end{theorem}
The $\alpha_1/\epsilon$ term represents the exploration cost—with higher accuracy (smaller $\epsilon$) requiring a larger upfront cost. Conversely, the $\alpha_2 \epsilon T$ term represents the pay-off during exploitation. This estimation-control trade-off is proven to be an unavoidable aspect of the problem itself, in Appendix \ref{proof:lower_bound}, establishing the $\mathcal{O}(T^{1/2})$ regret achieved by \scale{} in the full-rank setting (Theorem \ref{thm:main_guarantees}) as order-optimal.
\begin{algorithm}
    \caption{\hyscale{}}\label{alg:2h}
    \vspace{-4pt}
    \flushleft \textbf{Input:} noise cap $\bar{\eta}$, rank $r$, floor $\underline{\sigma_{r}^A}$, horizon $T$
    \begin{algorithmic}[1]
    \STATE Receive $(x_{m+1}^{\scale{}},\hat{A}^{\scale{}})$ from \scale{$(\bar{\eta},r,\underline{\sigma_{r}^A},T)$}
    \STATE $\hat{A}^{(m+1)} \gets \hat{A}^{\scale{}}$
    \STATE $x_{m+1} \gets x_{m+1}^{\scale{}}$
    \STATE $y_{m+1} \gets f_{m+1}(x_{m+1}) + \eta_{m+1}$
    \FOR{$t = m+2,\ldots,T$}
        \STATE $M^{(t)} \gets \hat{A}^{(t-1)} - \xi \cdot \nabla \ell_2\text{-loss}(y_{t-1},\hat{A}^{(t-1)})$
        \STATE $\hat{A}^{(t)} \gets \Pi_{M\succeq 0}\left(M^{(t)}\right)$
        \STATE $\{\hat{C}_k\}_{k=t}^T \gets $ RecurGen$(\hat{A}^{(t)},t)$
        \STATE $x_t \gets \hat{C}_t x_{t-1}+(I-\hat{C}_t)v_t$
        \STATE $y_{t} \gets f_{t}(x_{t}) + \eta_{t}$
    \ENDFOR
    \end{algorithmic}
\end{algorithm}

\subsection{\hyscale{}: Heuristic Adaptive Exploitation}
The strict explore-then-exploit structure of \scale{} intentionally discards all feedback received during the exploitation phase to preserve the $\epsilon$-estimation guarantee on $A$. This motivates a natural question: \textit{can we improve practical performance by additionally leveraging feedback during exploitation phase?}

To investigate this, we introduce \hyscale{} (Hybrid \scale{}), a heuristic algorithm that augments the exploitation phase of \scale{} with a continuous learning component. After the initial exploration and estimation, \hyscale{} uses the feedback from each subsequent round to refine its estimate of $A$. At round $t > m$, it performs a \textit{projected gradient descent} update on the previous estimate $\hat{A}^{(t-1)}$ using the noisy observed cost $y_{t-1}$ from the prior step:
$$
{\begin{split}
    A^{(t)} &=  \Pi_{M\succeq 0}\bigg\{\hat{A}^{(t-1)} - \xi (x_{t-1}-v_{t-1})(x_{t-1}-v_{t-1})^T \times \\
    \ldots & \times\left(\langle \hat{A}^{(t-1)}, (x_{t-1}-v_{t-1})(x_{t-1}-v_{t-1})^T \rangle - y_{t-1}\right)\bigg\}
\end{split}    }
$$
and then uses $A^{(t)}$ to re-compute $\hat{C}_t(\hat{A}^{(t)})$, for the current round, as detailed in Algorithm \ref{alg:2h}. As we will demonstrate next, this heuristic yields performance gains in light-tailed processes, where the feedback during exploitation proves to be a reliable signal. 
\begin{remark}\label{remark:hyscale_issues}
    Noisy gradient-based updates to \scale{}'s estimate $\hat{A}^{\scale{}}$ (from \eqref{eqn: trace_min}) can potentially push \hyscale{} into statistical inconsistency. We observe this in heavy-tailed processes in Figure \ref{fig:heavy_tails}.
\end{remark}

\section{Numerical Experiments}\label{sec:numerics}
To corroborate our theoretical findings and evaluate our proposed algorithms, we conduct a series of comprehensive numerical simulations against several baselines.
\subsection{Experimental Setup}
\vspace{-5pt}
Our simulation environment and the algorithms under consideration are configured as follows:
\vspace{-5pt}
\paragraph{Environment Generation.}
The online test environment is characterized by a general high-dimensional quadratic hitting cost $f_t(x_t) = \frac{1}{2}(x_t - v_t)^T A (x_t - v_t)$, a switching cost $c(x_t, x_{t-1}) = \frac{1}{2}\|x_t - x_{t-1}\|^2$ and the online sequence of its minimizers $\{v_t\}_{t=1}^T$.
\begin{figure}[]
    \centering
     \subfigure{%
        \includegraphics[width=0.32\linewidth]{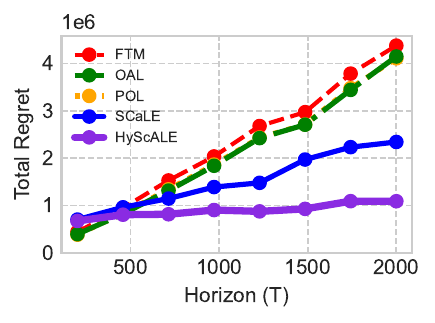}
        \label{fig:1a}}
     \subfigure{%
    \includegraphics[width=0.32\linewidth]{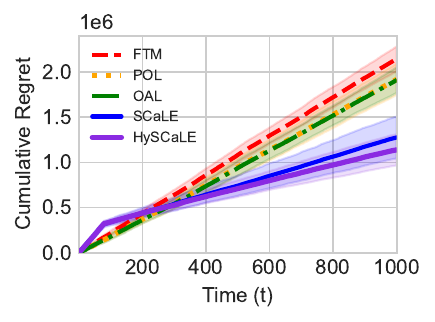}
    \label{fig:1b}}
     \subfigure{%
\includegraphics[width=0.32\linewidth]{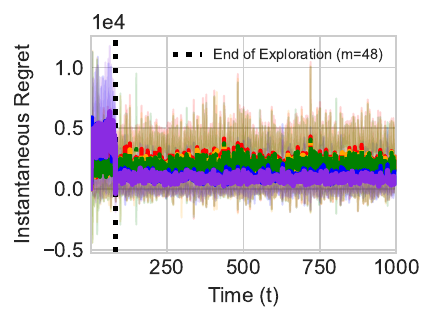}
    \label{fig:1c}}
    \caption{$\regret_T$, $\regret_{1000}(t)$ and $\frac{\Delta \regret_{1000}(t)}{\Delta t}$, for $r=d=4, \sigma_r^A = 10^{-2}, \sigma_v = 50,\bar{\eta} = 10$ and $c_1$ set to $3$}
    \label{fig:general_experiment}
\end{figure}
(a) \noindent \textbf{Cost Matrix $A$:} For precise control, we construct $A$ via spectral decomposition, $A = U S U^T$. $U$ is a randomly generated orthonormal matrix, and $S$ is a diagonal matrix of $r=rank(A)$ singular values. 

(b) \noindent \textbf{Minimizer Sequence $\{v_t\}_t$:} With $v_t = v_{t-1} +\delta_{t}$, we consider four cases for the underlying $\{\delta_t\}_t$-process: (i) uncorrelated light-tailed $(\delta_t\sim \text{ IID }\mathcal{N}(\mathbf{0},\sigma_v^2 I)$, (ii) correlated light-tailed $(\delta_t = (1-\alpha)\delta_{t-1}+\alpha\mathcal{N}(\mathbf{0},\sigma_v^2 I))$, (iii) heavy-tailed (Laplace distribution) and, (iv) fat-tailed (Cauchy distribution).


(c) \noindent \textbf{Feedback Noise:} Feedback $y_t$ has an additive noise $\eta_t$, drawn uniformly from $[-\bar{\eta}, \bar{\eta}]$. 

\paragraph{Algorithms and Baselines.}
We evaluate the performance of our two proposed algorithms against a theoretical optimum and three informative baselines.
\begin{itemize}
    \setlength\itemsep{0.5pt}
    \item Dynamic Online Optimal (\ai{}): Theoretical benchmark \cite{BhuyanMukherjee24} that has full knowledge of the matrix $A$, against which $\regret_T$ is measured.
    \item Follow the Minimizer (\ftm{}): The natural baseline in bandit feedback, where $x^{\ftm{}}_t = v_t.$ 
    \item Passive Online Learner (\pol{}): Designed to mimic online gradient descent. It starts with a naive estimate $\hat{A}_0 = I$ and at each round $t$:\\
    (i) it updates its estimate from $\hat{A}^{(t-1)}$ to $\hat{A}^{(t)}$ via a single projected gradient descent step on the observed \textit{noisy} feedback $y_t$.
    $$
        \hat{A}^{(t)} = \Pi_{M\succeq 0} \left\{\hat{A}^{(t-1)} - \xi \cdot \nabla \ell_2\text{-loss}\left(y_t,\hat{A}^{(t-1)}\right)\right\}
    $$ 
    (ii) re-computes $\{\hat{C}_k\}_{k=t}^T$ (using \texttt{RecurGen}$(\hat{A}^{(t)},t)$) and plays $x_t = \hat{C}_t x_{t-1}+(I-\hat{C}_t)v_t$, 
    \item Oracle-Assisted Learner (\oal{}): Powerful variant of \pol{} having access to \textbf{noiseless} feedback $y^*_t$ 
    \item \scale{} and \hyscale{}: Our Algorithms \ref{alg:2} and \ref{alg:2h}
\end{itemize}

\paragraph{Hyperparameter and Solver Settings.}
The learning rate $\xi$ for \hyscale{}, \pol{} and \oal{} is fixed at $\simeq 10^{-6}$.
The exploration parameter $c_1$, which defines the exploration length $m = c_1 r d$, is set to a fixed value $(\approx 3)$, upon which we further elaborate later in this section. All convex optimization problems are solved using the MOSEK solver \cite{MOSEK} (alternatively, an open-source Splitting Cone Solver (SCS) \cite{SCS}) via the CVXPY interface \cite{CVXPY}.

\subsection{Case Studies}\label{subsec:case_studies}
We present our key empirical findings here, as case studies, with all our results presented in Appendix \ref{appendix:addn_exps}.
\vspace{-5pt}
\paragraph{Primary Evaluation.}
We generate cumulative and instantaneous regret curves, averaged over multiple runs, for all algorithms. To verify the theoretical scaling of regret with the horizon (Theorem \ref{thm:main_guarantees}), we analyze $\regret_T$ as a function of $T \in [200,2000]$.

Figure \ref{fig:general_experiment}, where we consider \textbf{\textit{correlated}} $(\alpha=0.7)$ light-tailed process, shows that \scale{} and \hyscale{} significantly outperform all baselines, especially for longer horizons. Cumulative and Instantaneous regret reveal the initial high cost incurred by both \scale{} and \hyscale{} (up to $m=80$), which significantly lowers regret in the exploitation phase. Comparing \hyscale{} against \pol{} and \oal{} reveals that the benefits from projected GD are seen only after ensuring the baseline statistical estimate $\hat{A}^{\scale{}}$. 
\paragraph{Robustness to Heavy Tails.} Our goal is verifying Remark \ref{remark: Cai-Zhang} in the online setting:
\begin{quote}
    \textit{Is Restricted Uniform Boundedness (RUB) property of $\hat{A}$ necessary to ensure stability in bandit online quadratic optimization?}
\end{quote}
Specifically, the actions taken by \pol{} and \oal{}, of the form,
$$
    x_t = \hat{C}_t x_{t-1} + (I-\hat{C}_t)v_t
$$
are not centered sub-Gaussian. Consequently, the RUB property (and statistical consistency) cannot be guaranteed for these baselines. Likewise for \hyscale{} (see Remark \ref{remark:hyscale_issues}) due to noisy gradient updates.

\begin{figure}[]
    \centering
    \subfigure{%
    \includegraphics[width=0.23\linewidth]{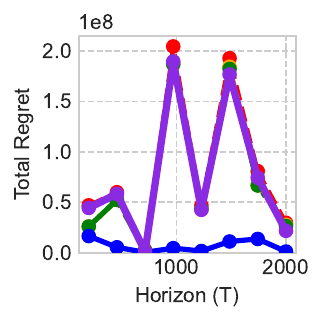}
    }
    \subfigure{%
    \includegraphics[width=0.23\linewidth]{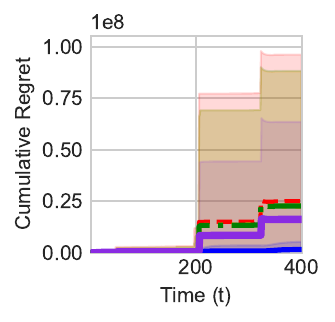}
    }
     \subfigure{%
        \includegraphics[width=0.23\linewidth]{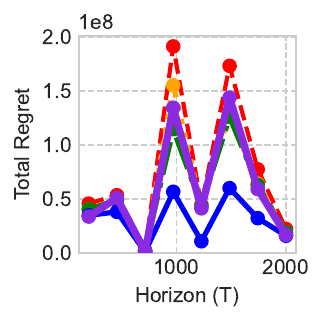}
        }
    \subfigure{%
    \includegraphics[width=0.23\linewidth]{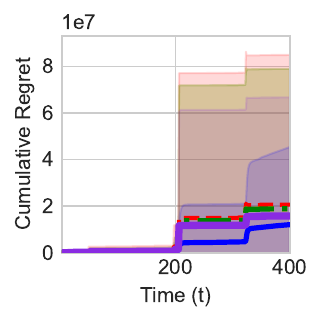}
    }\\    
    \vspace{-15pt}
    \subfigure{%
        \includegraphics[width=0.45\linewidth]{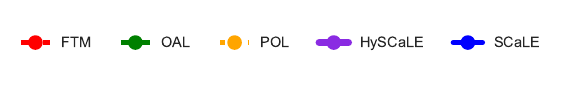}
        }
    \subfigure{%
        \includegraphics[width=0.45\linewidth]{figures/cumu_legend.pdf}
        }
    \vspace{-20pt}
    \caption{$v_t-v_{t-1} \sim$ Cauchy distribution with $\sigma_v = 1$. Left plots for $\sigma_r^A = 10^{-2}$ and right plots for $\sigma_r^A = 1$. Test settings: $r=1$, $d=4$, $\bar{\eta} = 50$, $c=10$.}
    \label{fig:heavy_tails}
\end{figure}

Figure \ref{fig:heavy_tails} shows gradient-based methods like \hyscale{} and the baselines \pol{} and \oal{} failing under when $\{v_t\}_t$ is governed by the Cauchy distribution. \scale{}, however, performs significantly better than the rest, especially when $\sigma^A_r$ is small. This illustrates the importance of exploration phase (Lines 1-5) and trace-norm minimization (Line 6) in Algorithm \ref{alg:2}.

\begin{figure}[]
    \centering
    \subfigure{%
    \includegraphics[width=0.23\linewidth]{figures/light_tail/horizon_regret_linear_d4_r1_v50_c110.0_runs5_sigma0.01_seed0.pdf}
    }
    \subfigure{%
        \includegraphics[width=0.23\linewidth]{figures/light_tail/cumulative_regret_T400_d4_r1_v50_c110.0_runs20_sigma0.01_seed0.pdf}
        }
     \subfigure{%
        \includegraphics[width=0.23\linewidth]{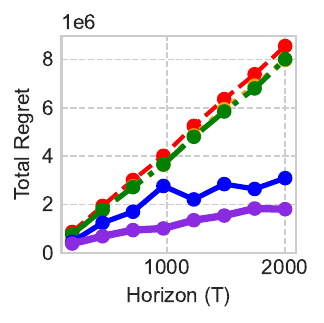}
        }
    \subfigure{%
    \includegraphics[width=0.23\linewidth]{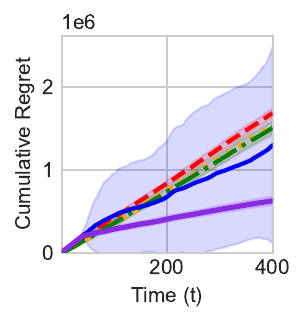}
    }\\
    \vspace{-15pt}
    \subfigure{%
        \includegraphics[width=0.45\linewidth]{figures/tot_legend.pdf}
    }
    \subfigure{%
        \includegraphics[width=0.45\linewidth]{figures/cumu_legend.pdf}
    }
    \vspace{-20pt}
    \caption{\scale{} and \hyscale{} for $r=1, d=4.$ Left plots for $\sigma_r^A = 10^{-2}$ and right plots for $\sigma_r^A = 1.$
    Test settings: $\bar{\eta} = 50$,$v_t-v_{t-1}\sim \mathcal{N}(0,50\cdot I)$, $c=10$}
    \label{fig:experiments-rank}
\end{figure}

\textbf{Rank-deficient Settings.} To test curvature signal $(A)$ recovery, we consider the setting where $r\ll d$ (recall Figure \ref{fig:3-d}). Figure \ref{fig:experiments-rank} depicts the performance of the considered algorithms when $r=1$ and $d=4.$
In this degenerate setting, the order difference claimed in Theorem \ref{thm:main_guarantees} is clearly visible in \scale{} and \hyscale{}'s dynamic regret.



\paragraph{Hyperparameter tuning.} Despite claimed universality of $c_1$, it is a hyperparameter in practical settings and requires tuning.
\begin{figure}[]
    \centering
    \subfigure{%
    \includegraphics[width=0.23\linewidth]{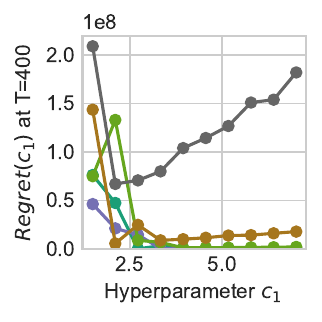}
    }
     \subfigure{%
        \includegraphics[width=0.23\linewidth]{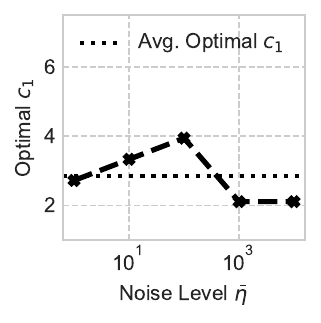}
        }\\
        \vspace{-15pt}
    \subfigure{%
        \includegraphics[width=0.45\linewidth]{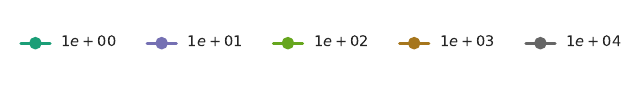}
        }
        \vspace{-20pt}
        \caption{$c_1$ v/s $\bar{\eta}$ sensitivity analysis at $T=400.$ Environment settings: $\sigma_r^A=1.0$, $r=4$, $d=4$, $\sigma_v = 50$}
    \label{fig:hyperparameter}
    \vspace{-10pt}
\end{figure}

We, therefore, set $c_1$ for each of the aforementioned experiments through a line-search that minimizes $\regret_T$:
\begin{align}\label{eqn:c1_search}
    c_1 = \argmin_{c \in \R^+} \regret_T(c).
\end{align}
Our sensitivity analysis reveals two key insights, illustrated in Figure \ref{fig:hyperparameter}. First, $\regret_T(c)$ exhibits a U-trend, pointing to an exploration-exploitation trade-off with respect to ``$m$''. Second, the optimal $c_1$ is observed to remain almost constant $(\approx 3)$ across a broad range of noise levels $\bar{\eta} \in [1,10^4]$, confirming that $c_1$ indeed is a universal parameter.

\section{Phase-wise Regret Analysis with Spectral Decomposition}
\subsection{Proof Sketch}\label{sec:proof_sketch}
This section presents the proofs for the main regret guarantees.
Our analysis proceeds by decomposing the total regret of \scale{} into contributions from the exploration and exploitation phases. It then employs spectral perturbation analysis to uncover that the distinct scaling of spectral errors as a function of estimation quality ($\|\hat{A}-A\| = \epsilon$) and the horizon ($T$) is precisely what accounts for the different regret behaviors

\begin{theorem}\label{thm:regret_decomp}
The regret of \scale{} is bounded as 
\begin{align}
        \begin{split}
    &\regret^{\text{\scale{}}}_T \leq \underbrace{m\left(\frac{(1/2)tr(\Sigma)}{1+\sigma^A_{\min}}\right)}_{\text{Irreducible Cost of NOT knowing } A} + \underbrace{m\gamma^2 \cdot \left((1/2)tr(A)+d\right)+ \frac{\gamma^2}{2}d}_{\text{Cost of Data Collection }\propto \frac{1}{\epsilon}}  + \underbrace{\regret_{m+1:T}^\scale{}}_{\substack{\text{Potential Benefit: }\\ o(T) \text{ Regret}}}
    \end{split}
    \end{align}
\end{theorem}

This decomposition makes the role of the signal-to-noise parameter, $\gamma^2$, explicit. Choosing $\gamma^2$ optimally requires balancing the second and third term, the practical effect of which is visible in our experiments in Figure~\ref{fig:general_experiment}(iii). We defer the proof of the above decomposition to Appendix \ref{proof:regret_decomp} to focus on the main analytical challenge: establishing a tight, non-trivial bound on $\regret_{m+1:T}^\scale{}$ as a function of the estimation error $\epsilon$.

We develop a novel spectral regret analysis, that further splits the exploitation regret into two orthogonal components: one driven by errors in the estimated eigenvalues of $A$, and the other by perturbations in its estimated eigenbasis. Define the class of algorithms $\ai{(\hat{A})}: \{M\succeq \boldsymbol{0}_{d\times d}\}\to \R^{d\times T}$, using the structure of the online optimal \ai{} benchmark and estimate $\hat{A}$:
\begin{align}
    \{\hat{C}_t\}_{t=1}^T &\gets \texttt{RecurGen}(\hat{A},1)\\
    x^{\ai{(\hat{A})}}_{t} &= \hat{C}_t x^{\ai{(\hat{A})}}_{t-1}+(I-\hat{C}_t)v_t \text{ } \forall \text{ } t
\end{align}
Let the eigenvalue decompositions be $A = QDQ^T$ and $\hat{A} = P\hat{D}P^T$.
\begin{proposition}\label{prop:spectral_split}
For $\{v_t\}_{t=1}^T$ as any martingale sequence, the dynamic regret of \ai{$(\hat{A})$} is precisely
\begin{align}
    \begin{split}
        \regret_T^{\ai{(\hat{A})}} =& \underbrace{\sum_{s=1}^{T} \frac{1}{2}\E\|v_{s}-v_{s-1}\|_{C_{s}-\hat{C}_{s}}^2}_{R_T^{val}: \text{ dominated by }\Delta_{val} = \hat{D}-D} +\underbrace{\sum_{s=1}^{T} \frac{1}{2}\E\|x_{s-1}-v_s\|_{\hat{C}_s(A-\hat{A})\hat{C}_s}^2}_{R_T^{basis}: \text{ dominated by } \Delta_{basis} = P-Q}
    \end{split}
    \end{align}
\end{proposition}
This decomposition, proved in Appendix \ref{proof:spectral_split}, is a non-trivial extension of the analysis done in \cite{BhuyanMukherjee24}; decoupling spectral analyses across the horizon $\{1,\ldots,T\}$ to generate individual regret terms $R_T^{val}$ and $R_T^{basis}.$

The next challenge is a tight translation of nuclear norm error $\|\hat{A}-A\|_*$ into the two regret components:
\begin{center}
    \begin{tikzpicture}
    \node(formula){$\|\hat{A}-A\|_*\leq \epsilon$};
    \node(solution1) [above right =-0.5em and 2em of formula]{$\Delta_{val}(\epsilon) \longrightarrow R_T^{val}(\epsilon,T)$};
    \node(solution2) [below right =-0.5em and 2em of formula]{$\Delta_{basis}(\epsilon) \longrightarrow R_T^{basis}(\epsilon,T)$};
    \draw [->] (formula.east) to [out=0, in=180] (solution1.west);
    \draw [->] (formula.east) to [out=0, in=180] (solution2.west);
\end{tikzpicture}
\end{center}
Matrix perturbation theory \cite{cai2018rate} is the key tool here. A central finding of our work is that $R_T^{val}(\epsilon,T)$ and $R_T^{basis}(\epsilon,T)$ contribute equally to the total regret for both cases of rank-deficient $(r<d)$ and full-rank $A.$

We begin with the more challenging rank-deficient case. The presence of singular regions in $A$ prevents the ``decaying error propagation'' that standard control/regret analyses \cite{9483309,10590739} rely on. Naive analysis lets estimation errors accumulate linearly $(\simeq \epsilon t)$ at each round~$t$, leading to an uncontrollable, total regret of $\epsilon T^2$.  We overcome this by tightly characterizing the error accumulation from eigenbasis perturbation: 

\begin{theorem}\label{thm:exploit_regret_degen}
Let $rank(A) = r < d$. For any estimate $\hat{A}$ such that $\|\hat{A}-A\|_*\leq \epsilon$, the dynamic regret of \ai{$(\hat{A})$} is bounded by
\begin{align}
    \begin{split}
        \regret_T^{\ai{(\hat{A})}} \leq& \underbrace{\left(\frac{2d\sigma^A_{\max} \|\Sigma\|_F}{(\sigma^A_{\min}-\epsilon)^2}\right)\epsilon^2 T^2}_{R_T^{val}} + \underbrace{tr(\Sigma)\sqrt{\epsilon}T}_{R_T^{basis}}+\epsilon T \cdot \frac{(3\sqrt{2d}+2\epsilon d + 2\sigma^A_{\max}\sqrt{2d})\|\Sigma\|_F}{(\sigma^A_{\min}-\epsilon)^2}
        \end{split}
\end{align}
\end{theorem}
Crucially, the dominant error term, arising from eigenbasis perturbation, is controlled not by $\epsilon$, but by a much smaller $\epsilon^2$ factor. Over the horizon, this error evolves into $\mathcal{O}(\epsilon^2 T^2)$ regret, instead of (uncontrollable) $\mathcal{O}(\epsilon T^2)$. Our dissected spectral analysis, that reveals this structure, is laid out in the coming Section \ref{proof:exploit_regret_degen}.

When $A$ is full-rank, the error at each step is a \textit{constant}, owing to geometrically decaying error propagation through the horizon. Applying the analysis developed for the rank-deficient case (Section \ref{proof:exploit_regret_degen}), leads to the following improved regret guarantee, details of which are in Appendix \ref{proof:exploit_regret_full}.
\begin{theorem}\label{thm:exploit_regret_full}
Let $rank(A) = d$. For any estimate $\hat{A}$ such that $\|\hat{A}-A\|_*\leq \epsilon$, the exploitation regret of \ai{$(\hat{A})$} is bounded by
$$
    \regret_T^{\ai{(\hat{A})}} \leq \underbrace{\epsilon T\cdot \frac{(1+(\sqrt{2}+\epsilon\sqrt{d})\sigma^A_{\max})\sqrt{d} \|\Sigma\|_F }{(\sigma^A_{\min}-\epsilon)^3}}_{R_T^{val}}+ \underbrace{\epsilon T \cdot \frac{tr(\Sigma)}{2\sigma^A_{\min}}}_{R_T^{basis}}.
$$
\end{theorem}
Finally, to achieve an estimation error of $\|\hat{A}-A\|_* \leq \epsilon$ via rank-1 measurements, the exploration phase must incur a cost of $\Theta(1/\epsilon)$. The following corollary shows that balancing this cost against the exploitation regret from Theorems \ref{thm:exploit_regret_degen} and \ref{thm:exploit_regret_full} leads to an equal contribution from both spectral error terms:
\begin{corollary}\label{corr:equal-contribution}
The optimal balance between the exploration regret, $\regret_{1:m} = \Theta(1/\epsilon)$, and the exploitation regret, $\regret_{m+1:T}^{(\alg{\hat{A}})}$, is achieved:
\begin{enumerate}
    \item for rank-deficient $(r<d)$ $A$ when
        $$R_T^{val}=\mathcal{O}(T^{2/3}) \quad \text{ and } \quad R_T^{basis}=\mathcal{O}(T^{2/3}),$$
    \item for full-rank $(r=d)$ $A$ when 
        $$R_T^{val}=\mathcal{O}(T^{1/2}) \quad \text{ and } \quad R_T^{basis}=\mathcal{O}(T^{1/2}).$$
\end{enumerate}
\end{corollary}
The detailed proof in Appendix \ref{proof:equal-contribution} gives the exact regret expressions and the optimal value of $\gamma^2$ used in \scale{} (Algorithm \ref{alg:2}). It links the detailed analysis in this section back to the main guarantee presented in Theorem \ref{thm:main_guarantees}.
The key insight from our spectral decomposition is that explore-exploit balance is achieved when regret contributions from eigenvalue errors ($R_T^{val}$) and eigenbasis errors ($R_T^{basis}$) are of the same order.

\subsection{Main Analytical Contribution: Proof of Theorem \ref{thm:exploit_regret_degen}}\label{proof:exploit_regret_degen}
This section highlights the key methodological contribution of this work: sub-linear regret under growing errors, typically arising from the degeneracy in curvature of the objective function. To that end, we directly utilize standard matrix perturbation results (refer to Appendix \ref{appendix:matrix_perturb_basics}) and Proposition \ref{prop:spectral_split} (proof in Appendix \ref{proof:spectral_split}). Now, lets again go back to the exact generalized regret formulation we have:
\begin{align}
    \text{Regret}[1,T] = \sum_{s=1}^{T} \frac{1}{2}\E\|x_{s-1}-v_{s}\|_{\hat{C}_{s} (A-\hat{A}) \hat{C}_{s}}^2 + \sum_{s=1}^{T} \frac{1}{2}\E\|v_{s-1}-v_{s}\|_{C_s-\hat{C}_{s}}^2
\end{align}
where
\begin{align}
    \E\|x_{s-1}-v_{s}\|_{\hat{C}_{s} (A-\hat{A}) \hat{C}_{s}}^2 = \sum_{k=0}^{s-1} \E\|v_{s-k-1}-v_{s-k}\|_{\prod_{l=k}^0\hat{C}_{s-l} (A-\hat{A}) \prod_{l=0}^k\hat{C}_{s-l}}^2.
\end{align}
We look at
\begin{align}
    &\prod_{l=k}^0\hat{C}_{s-l} (A-\hat{A}) \prod_{l=0}^k\hat{C}_{s-l} \\
    =& P\times diag\left(\prod_{l=k}^0 \hat{\lambda}^1_{s-l},\ldots,\prod_{l=k}^0 \hat{\lambda}^d_{s-l}\right)\times \underbrace{(P^T Q D Q^T P - \hat{D})}_{P^T Q D Q^T P - D + D - \hat{D}} \times diag\left( \prod_{l=k}^0 \hat{\lambda}^1_{s-l},\ldots,\prod_{l=k}^0 \hat{\lambda}^d_{s-l}\right) \times P^T.\\
    =& \underbrace{P\times diag\left(\prod_{l=k}^0 \hat{\lambda}^1_{s-l},\ldots,\prod_{l=k}^0 \hat{\lambda}^d_{s-l}\right)\times \underbrace{(P^T Q D Q^T P - D)}_{\Delta_2} \times diag\left(\prod_{l=k}^0 \hat{\lambda}^1_{s-l},\ldots,\prod_{l=k}^0 \hat{\lambda}^d_{s-l}\right) P^T}_{X_1}\\
    &+  \underbrace{P\times diag\left(\prod_{l=k}^0 \hat{\lambda}^1_{s-l},\ldots,\prod_{l=k}^0 \hat{\lambda}^d_{s-l}\right)\times (D-\hat{D}) \times diag\left(\prod_{l=k}^0 \hat{\lambda}^1_{s-l},\ldots,\prod_{l=k}^0 \hat{\lambda}^d_{s-l}\right) P^T}_{X_2}.
\end{align}
Lets look at $X_2$ first:
\begin{align}
    D - \hat{D} &= diag(\lambda^{A,1}-\hat{\lambda}^1,\ldots,\lambda^{A,d}-\hat{\lambda}^d)\\
    &\preceq diag(\underbrace{\epsilon,\ldots,\epsilon}_{r},\underbrace{0,\ldots,0}_{d-r})
\end{align}
Also, the following holds
\begin{align}
    \hat{\lambda}^i_t \leq \begin{cases}
        \frac{1}{1+\lambda^{A,i}-\epsilon} & i \in \{1,\ldots,r\}\\
        1 & i\in \{r+1,\ldots,d\}
    \end{cases}
\end{align}
which will give us
\begin{align}
    &diag\left(\prod_{l=k}^0 \hat{\lambda}^1_{s-l},\ldots,\prod_{l=k}^0 \hat{\lambda}^d_{s-l}\right) \times (D - \hat{D}) \times diag\left(\prod_{l=k}^0 \hat{\lambda}^1_{s-l},\ldots,\prod_{l=k}^0 \hat{\lambda}^d_{s-l}\right)\\
    &\preceq \underbrace{diag\left(\frac{\epsilon}{(1+\lambda^{A,1}-\epsilon)^{2k+2}},\ldots,\frac{\epsilon}{(1+\lambda^{A,r}-\epsilon)^{2k+2}},0,\ldots,0\right)}_{\Delta_1}
\end{align}
leading to
\begin{align}
    \E\|v_{s-k-1}-v_{s-k}\|_{X_2}^2 &= \langle X_2,\Sigma\rangle\\
    &= \langle \Delta_1, P^T\Sigma P\rangle\\
    &= \sum_{i=1}^r \frac{\epsilon}{(1+\lambda^{A,i}-\epsilon)^{2k+2}} p_i^T \Sigma p_i\\
    &\leq \frac{\epsilon}{(1+\sigma^A_{\min}-\epsilon)^{2k+2}}tr(P^T\Sigma P)\\
    &= \epsilon \cdot \frac{tr(\Sigma)}{(1+\sigma^A_{\min}-\epsilon)^{2k+2}}
\end{align}
We will now look at $X_1$, and more specifically, $\Delta_2$:
\begin{align}
    \Delta_2 &= P^T Q D Q^T P - D\\
    &= (P^T Q-I + I) D (Q^T P -I +I) -D\\
    &= (P^T Q-I)D (Q^T P -I) + (P^T Q-I)D + D (Q^T P -I) + D-D\\
    &= \underbrace{(P^T Q-I)D (Q^T P -I)}_{\Delta_{2a}} + \underbrace{(P^T Q-I)D}_{\Delta_{2b}} + \underbrace{D (Q^T P -I)}_{\Delta_{2c}}
\end{align}
Now, observe that
\begin{align}
     \Delta_{3b} &= diag\left(\prod_{l=k}^0 \hat{\lambda}^1_{s-l},\ldots,\prod_{l=k}^0 \hat{\lambda}^d_{s-l}\right) \times \Delta_{2b}  \times diag\left(\prod_{l=k}^0 \hat{\lambda}^1_{s-l},\ldots,\prod_{l=k}^0 \hat{\lambda}^d_{s-l}\right)\\
     &= diag\left(\prod_{l=k}^0 \hat{\lambda}^1_{s-l},\ldots,\prod_{l=k}^0 \hat{\lambda}^d_{s-l}\right) \times (P^TQ-I) \times diag\left(\underbrace{\lambda^{A,1}\prod_{l=k}^0 \hat{\lambda}^1_{s-l},\ldots,\lambda^{A,r}\prod_{l=k}^0 \hat{\lambda}^r_{s-l}}_{r},\underbrace{0,\ldots,0}_{d-r}\right) 
\end{align}
and
\begin{align}
     \Delta_{3c} &=diag\left(\prod_{l=k}^0 \hat{\lambda}^1_{s-l},\ldots,\prod_{l=k}^0 \hat{\lambda}^d_{s-l}\right) \times \Delta_{2c}  \times diag\left(\prod_{l=k}^0 \hat{\lambda}^1_{s-l},\ldots,\prod_{l=k}^0 \hat{\lambda}^d_{s-l}\right)\\
     &= diag\left(\underbrace{\lambda^{A,1}\prod_{l=k}^0 \hat{\lambda}^1_{s-l},\ldots,\lambda^{A,r}\prod_{l=k}^0 \hat{\lambda}^r_{s-l}}_{r},\underbrace{0,\ldots,0}_{d-r}\right) \times (P^TQ-I)  diag\left(\prod_{l=k}^0 \hat{\lambda}^1_{s-l},\ldots,\prod_{l=k}^0 \hat{\lambda}^d_{s-l}\right).  
\end{align}
Lastly, $\Delta_{3a}$ is defined accordingly,
\begin{align}
    \Delta_{3a} &=diag\left(\prod_{l=k}^0 \hat{\lambda}^1_{s-l},\ldots,\prod_{l=k}^0 \hat{\lambda}^d_{s-l}\right) \times \Delta_{2a}  \times diag\left(\prod_{l=k}^0 \hat{\lambda}^1_{s-l},\ldots,\prod_{l=k}^0 \hat{\lambda}^d_{s-l}\right).
\end{align}
Putting these three together now
\begin{align}
     \E\|v_{s-k-1}-v_{s-k}\|_{X_1}^2 &= \langle \Delta_{3a},\P^T \Sigma P\rangle + \langle \Delta_{3b},\P^T \Sigma P\rangle + \langle \Delta_{3c},P^T \Sigma P\rangle\\
     &\leq \|\Sigma\|_F\cdot (\|\Delta_{3a}\|_F+\|\Delta_{3b}\|_F+\|\Delta_{3c}\|_F)
\end{align}
Now, $\|P^TQ-I\|_F = \|Q^TP-I\|_F = \|P-Q\|_F\leq \sqrt{d}\cdot\|P-Q\|_{op}\leq \frac{\sqrt{2d}\cdot \epsilon}{\sigma^A_{\min}-\epsilon}$. We, therefore, have
\begin{align}
    \|\Delta_{3b}\|_F = \|\Delta_{3c}\|_F \leq \frac{\sigma^A_{\max} \sqrt{2d} \cdot\epsilon}{(\sigma^A_{\min}-\epsilon)(1+\sigma^A_{\min}-\epsilon)^{k+1}}
\end{align}
and
\begin{align}
    \|\Delta_{3a}\|_F &\leq \|\Delta_{2a}\|_F\\
    &\leq \|P^TQ-I\|_F \cdot \|D(Q^TP-I)\|_F\\
    &\leq  \frac{2d\sigma^A_{\max} \cdot \epsilon^2}{(\sigma^A_{\min}-\epsilon)^2}.
\end{align}
Now putting $X_1$ and $X_2$ together,
\begin{align}
    &\E\|x_{s-1}-v_{s}\|_{\hat{C}_{s} (A-\hat{A}) \hat{C}_{s}}^2 \\
    &\leq \sum_{k=0}^{s-1} \frac{2d\sigma^A_{\max} \|\Sigma\|_F \cdot \epsilon^2}{(\sigma^A_{\min}-\epsilon)^2} + 2\sum_{k=0}^{s-1} \frac{\|\Sigma\|_F\sigma^A_{\max} \sqrt{2d} \cdot\epsilon}{(\sigma^A_{\min}-\epsilon)(1+\sigma^A_{\min}-\epsilon)^{k+1}} +\sum_{k=0}^{s-1} \epsilon \cdot \frac{tr(\Sigma)}{(1+\sigma^A_{\min}-\epsilon)^{2k+2}}\\
    &\leq \left(\frac{2d\sigma^A_{\max} \|\Sigma\|_F}{(\sigma^A_{\min}-\epsilon)^2}\right)\epsilon^2\cdot s+\frac{2\|\Sigma\|_F\sigma^A_{\max} \sqrt{2d} \cdot\epsilon}{(\sigma^A_{\min}-\epsilon)(1+\sigma^A_{\min}-\epsilon)}+\frac{tr(\Sigma) \epsilon}{(1+\sigma^A_{\min}-\epsilon)^2-1}
\end{align}
which leads to a regret of 
\begin{align}
\begin{split}
    \text{Regret}[1,T] \leq& \left(\frac{d\sigma^A_{\max} \|\Sigma\|_F}{(\sigma^A_{\min}-\epsilon)^2}\right)\epsilon^2\cdot T(T+1)+\frac{2\|\Sigma\|_F\sigma^A_{\max} \sqrt{2d} \cdot\epsilon T}{(\sigma^A_{\min}-\epsilon)(1+\sigma^A_{\min}-\epsilon)}+\frac{tr(\Sigma) \epsilon T}{(1+\sigma^A_{\min}-\epsilon)^2-1} \\
    &+ \sum_{s=1}^{T} \frac{1}{2}\E\|v_{s-1}-v_{s}\|_{C_s-\hat{C}_{s}}^2.
\end{split}
\end{align}
We will now look at the term $\sum_{s=1}^{T} \frac{1}{2}\E\|v_{s-1}-v_{s}\|_{C_s-\hat{C}_{s}}^2$:
\begin{align}
    C_s - \hat{C_s} &= QD_sQ^T - P\hat{D}_sP^T\\
    &= P\left(P^T D_sQ^T P - \hat{D}_s\right)P^T
\end{align}
where
\begin{align}
    P^T D_sQ^T P - \hat{D}_s = \underbrace{(P^TQ -I)D_s(Q^T P- I)}_{\Delta_{4a}} +\underbrace{(P^TQ -I)D_s}_{\Delta_{4b}} + \underbrace{D_s(Q^T P- I)}_{\Delta_{4c}} + \underbrace{D_s-\hat{D}_s}_{\Delta_{4d}}
\end{align}
From our previous calculations:
\begin{align}
    \|\Delta_{4a}\|_F &\leq \|P^TQ -I\|_F \cdot \|D_s(P^TQ -I)\|_F\\
    &\leq \frac{\sqrt{2d}\cdot \epsilon}{\sigma^A_{\min}-\epsilon} \cdot 1 \cdot \frac{\sqrt{2d}\cdot \epsilon}{\sigma^A_{\min}-\epsilon}\\
    &\leq \frac{2d\epsilon^2}{(\sigma^A_{\min}-\epsilon)^2}\\
    \|\Delta_{4b}\|_F = \|\Delta_{4c}\|_F &\leq \frac{\sqrt{2d}\cdot \epsilon}{\sigma^A_{\min}-\epsilon}.
\end{align}
where we use the fact that each diagonal entry of $D_s$ is at most $1.$  We now focus on the last term:
\begin{align}
    D_s - \hat{D_s} = diag(\lambda^1_t-\hat{\lambda}^1_t,\ldots, \lambda^1_r-\hat{\lambda}^r_t,\lambda^{r+1}_t-\hat{\lambda}^{r+1}_t,\ldots,\lambda^d_t-\hat{\lambda}^d_t).
\end{align}
For $i\in\{1,\ldots,r\}$,
\begin{align}
    \frac{1}{\lambda^i_t} &= 2+\lambda^{A,i}-\lambda^i_{t+1}\\
    \frac{1}{\hat\lambda^i_t} &= 2+\hat\lambda^{A,i}-\hat\lambda^i_{t+1}\\
    \implies |\lambda^i_t-\hat{\lambda}^i_t| &\leq \lambda^i_t\hat{\lambda}^i_t\cdot |\lambda^i_{t+1}-\hat\lambda^i_{t+1}| +  \lambda^i_t\hat{\lambda}^i_t\cdot|\lambda^{A,i}-\hat\lambda^{A,i}|\\
    &\leq \frac{|\lambda^i_{t+1}-\hat\lambda^i_{t+1}|}{1+\lambda^{A,i}}  + \frac{|\lambda^{A,i}-\hat\lambda^{A,i}|}{1+\lambda^{A,i}}.
\end{align}
Doing this recursively gets
\begin{align}
    |\lambda^i_t-\hat{\lambda}^i_t| \leq \frac{|\lambda^{A,i}-\hat\lambda^{A,i}|}{\lambda^{A,i}} \leq \frac{\epsilon}{\min_{i\leq r}\lambda^{A,i}} =  \frac{\epsilon}{\sigma^A_{\min}}
\end{align}
for $i\in \{1,\ldots\}.$ Now, the only reason this holds for these indices is because $\lambda^{A,i}=0$ for $i>r.$ For these indices, $\lambda^i_t = 1$ for all $t.$ We therefore look at only the equation for $\hat{\lambda}^i_t$
\begin{align}
    \frac{1}{\hat\lambda^i_t} &= 2+\hat\lambda^{A,i}-\hat\lambda^i_{t+1}
\end{align}
where we know that $0\leq \hat\lambda^{A,i} < \epsilon$ for all $i>r.$ Stochastic dynamic programming analysis from \cite{BhuyanMukherjee24} shows that
\begin{align}
    \frac{\hat\lambda^{A,i} + 2 -\sqrt{(\hat\lambda^{A,i})^2+4\hat\lambda^{A,i}}}{2}=\hat{\lambda}^i_L \leq \hat{\lambda}^i_t < \frac{1}{1+\hat\lambda^{A,i}}
\end{align}
and 
\begin{align}
    \frac{\hat\lambda^{A,i} + 2 -\sqrt{(\hat\lambda^{A,i})^2+4\hat\lambda^{A,i}}}{2} > \frac{\epsilon+2-\sqrt{\epsilon^2+4\epsilon}}{2} = 1-\sqrt{\epsilon}+\frac{\epsilon}{2}>1-\sqrt{\epsilon}
\end{align}
giving us
\begin{align}
    \lambda^i_t - \hat{\lambda}^i_t \leq \sqrt{\epsilon} \text{ } \forall \text{ } i>r
\end{align}
We therefore have
\begin{align}
    (D_s-\hat{D}_s)_{i,i} \leq \max\left\{\sqrt{\epsilon},\frac{\epsilon}{\sigma^A_{\min}}\right\}
\end{align}
which we use in $\E\|v_{s-1}-v_{s}\|_{C_s-\hat{C}_{s}}^2$ to get
\begin{align}
    \E\|v_{s-1}-v_{s}\|_{C_s-\hat{C}_{s}}^2 \leq \left(\frac{2d\epsilon^2}{(\sigma^A_{\min}-\epsilon)^2} + \frac{\sqrt{2d}\cdot \epsilon}{\sigma^A_{\min}-\epsilon} \right)\|\Sigma\|_F + \max\left\{\sqrt{\epsilon},\frac{\epsilon}{\sigma^A_{\min}}\right\}\cdot tr(\Sigma). 
\end{align}
The final regret upper bound for an estimate $\hat{A}$ with error rate $\|\hat{A}-A\|_* \leq \epsilon$ is 
\begin{align}
    \begin{split}
        \text{Regret}_T \leq& \underbrace{\left(\frac{d\sigma^A_{\max} \|\Sigma\|_F}{(\sigma^A_{\min}-\epsilon)^2}\right)\epsilon^2\cdot T(T+1)+\frac{2\|\Sigma\|_F\sigma^A_{\max} \sqrt{2d} \cdot\epsilon T}{(\sigma^A_{\min}-\epsilon)(1+\sigma^A_{\min}-\epsilon)}+\frac{tr(\Sigma) \epsilon T}{(1+\sigma^A_{\min}-\epsilon)^2-1}}_{\text{Regret}_T^A} \\
        &+ \underbrace{T\left(\left[\frac{2d\epsilon^2}{(\sigma^A_{\min}-\epsilon)^2} + \frac{\sqrt{2d}\cdot \epsilon}{\sigma^A_{\min}-\epsilon} \right]\|\Sigma\|_F + \max\left\{\sqrt{\epsilon},\frac{\epsilon}{\sigma^A_{\min}}\right\}\cdot tr(\Sigma)\right)}_{\text{Regret}_T^B}. 
        \end{split}
\end{align}
Combining the above expression with the regret of the exploration phase, we get the following dynamics, that summarize the total regret, and our choice of $1/\gamma^2 = \Theta(\epsilon) \propto T^{-2/3}$ for $r<d$ in Algorithm \ref{alg:2}:
\begin{align}
    \text{Regret}_T^{\scale{}} = \mathcal{O}(\gamma^2) +  \mathcal{O}(\epsilon^2 T^2) + \mathcal{O}(\epsilon^{1/2}\cdot T).
\end{align}
The proof of Theorem \ref{thm:exploit_regret_full} and the consequent choice of $\gamma^2$ for the full-rank ($r=d$) case follows a similar technique, and can be found in Appendix \ref{proof:exploit_regret_full} and \ref{proof:equal-contribution}.

\section{Motivating Applications}\label{sec:applications}

In this section, we detail two real-world control problems---Data Center Thermal Management and Cooperative Wind Farm Control---that directly motivate our formulation. In both settings, the controller must optimize a quadratic objective governed by an unknown physical coupling matrix while tracking a stochastic minimizer, all under the constraint of significant switching costs.

\subsection{Data Center Thermal Management (Zonal Airflow Control)}
Modern data centers consume vast amounts of energy, a significant portion of which is dedicated to cooling infrastructure. The control objective is to modulate the airflow rates of Computer Room Air Conditioning (CRAC) units, denoted by $x_t \in \mathbb{R}^d$, to maintain safe server temperatures while minimizing energy consumption \cite{lazic2018data}.

\begin{itemize}
    \item \textbf{The Minimizer ($v_t$):} The optimal airflow configuration is dictated by the IT load (heat generation) distributed across server racks. As computational jobs follow stochastic arrival processes—often modeled as random walks or bursty Brownian motion—the ``target'' cooling profile $v_t$ behaves as a martingale process observed via CPU utilization metrics prior to actuation \cite{meisner2009power}.
    
    \item \textbf{The Unknown Matrix ($A$):} The matrix $A \succeq 0$ represents the \textit{convective thermal coupling} (or Heat Cross-Interference Matrix) between CRAC units and server racks, where $A_{ij}$ quantifies the cooling influence of fan $i$ on rack $j$ \cite{tang2007thermal,moore2006weatherman}. Despite facility ownership, this matrix is effectively unknown and typically time-varying due to complex turbulent airflow, recirculation, and physical degradation (e.g., clogged filters, cable displacement). Consequently, static offline models based on Computational Fluid Dynamics (CFD) often fail to capture real-time dynamics, necessitating online identification \cite{evans2016deepmind,gamble2018safety}.
    
    \item \textbf{The Cost Structure:} The hitting cost $f_t(x_t)$ models the trade-off between power usage and thermal risk. Deviation from the optimal airflow $v_t$ results in either cubic power penalties (over-cooling) or exponential hardware failure risk (under-cooling), standardly approximated via quadratic penalty functions \cite{zanini2013online}. Crucially, the system imposes a \textit{switching cost} proportional to $\|x_t - x_{t-1}\|^2$. Rapid fluctuation of fan speeds significantly degrades Variable Frequency Drives (VFDs) and induces acoustic fatigue, necessitating smooth control trajectories to maximize equipment lifetime.
\end{itemize}

\subsection{Cooperative Wind Farm Control (Wake Steering)}
In dense wind farms, upstream turbines generate ``wakes''—regions of turbulent, low-velocity air—that significantly reduce the power output of downstream turbines. ``Wake steering'' is a cooperative control strategy where upstream turbines intentionally misalign their yaw angles, denoted by $x_t \in \mathbb{R}^d$, to deflect wakes away from downstream neighbors, thereby maximizing total farm power \cite{fleming2017field,gebraad2016wind}.

\begin{itemize}
    \item \textbf{The Minimizer ($v_t$):} The optimal yaw configuration is strictly dependent on the instantaneous free-stream wind direction. Wind direction is a stochastic environmental variable that exhibits non-stationary drift, effectively acting as a martingale driving force observed by nacelle-mounted anemometers \cite{pinson2013wind,jiang2017data}.
    
    \item \textbf{The Unknown Matrix ($A$):} The matrix $A$ captures the \textit{aerodynamic interaction} (or sensitivity Jacobian) between turbines. It dictates how a change in the yaw angle of turbine $i$ impacts the power capture of downstream turbine $j$. While analytic surrogates like the Jensen or Gaussian wake models exist \cite{jensen1983note,bastankhah2014new}, the true interaction depends on unobservable atmospheric parameters such as turbulence intensity, shear, and thermal stability \cite{howland2019wind}. As these atmospheric conditions shift, the effective coupling $A$ changes, requiring the controller to continuously estimate local aerodynamic sensitivities \cite{ebegbulem2017distributed}.
    
    \item \textbf{The Cost Structure:} The objective is to maximize total power, which is locally convex near the operating point. The hitting cost $(x_t - v_t)^\top A (x_t - v_t)$ penalizes deviations from the optimal wake-deflection angles. The \textit{switching cost} is of paramount importance in this domain. Yawing the massive nacelle structure induces large gyroscopic loads that cause structural fatigue in the tower and blades (Damage Equivalent Loads) \cite{bossanyi2018wind}. To prevent mechanical failure and extend asset life, controllers must penalize the yaw rate through $\|x_t - x_{t-1}\|_2^2$ \cite{kanev2018active}.
\end{itemize}

\section{Related Works}\label{appendix_sec:lit_review}
This section provides a detailed overview of the research areas that intersect with our work, namely online algorithms, control theory, online learning, and inverse optimal control.

\subsection{Online Algorithms with Switching Costs}

This body of literature most directly addresses the challenge of managing movement costs in online decision-making. It is primarily composed of three related subfields: Metrical Task Systems (MTS), Convex Body Chasing (CBC), and Smoothed Online Convex Optimization (SOCO).

\paragraph{Metrical Task Systems (MTS)} Introduced as a foundational model for online problems with switching costs, MTS frames the problem as an online player moving a server on a metric space to serve a sequence of requests (tasks) \cite{bartal1997polylog,blum2000line}. The goal is to minimize the sum of movement costs and task-processing costs. The field has a long history of developing competitive algorithms and understanding the fundamental limits of the problem \cite{borodinOptimalOnlineAlgorithm1992,albers2003online,ChristiansonWierman23}, with strong applications in the domains of paging \cite{JPS20,BCKPV20}, k-server problems \cite{fiat1994competitive,Koutsoupias95} and caching \cite{lykouris2018competitive,LV21}, to name a few. 
Recent literature has shifted its focus to the use of machine-learned advice to provide robust yet opportunistic algorithms \cite{AGKK20,ChristiansonWierman23}.

\paragraph{Convex Body Chasing (CBC)} CBC can be viewed as a continuous and geometric generalization of MTS, where the player must maintain a point inside a sequence of convex sets (bodies) revealed online, minimizing the total distance moved \cite{Antoniadis18}. This framework has led to significant theoretical advances and connections to diverse areas of optimization \cite{sellkeChasingConvexBodies2020,argueChasingConvexBodies2021}. However, negative results have shown that the problem can be provably hard in high dimensions \cite{Antoniadis18}. 

\paragraph{Smoothed Online Convex Optimization (SOCO)} SOCO provides a functional optimization perspective of both MTS and CBC, defining the problem as minimizing the sum of a sequence of hitting cost functions and the switching costs between consecutive actions \cite{BansalGupta15,GoelLinWierman19,BhuyanMukherjee24,bhuyan2024optimal,bhuyan2025estimate}. This framework is particularly relevant to our work and has been successfully applied in domains involving (possibly) unbounded decision spaces, with data-center management \cite{LinWierman12}, adaptive control \cite{lin2021perturbation,GoelWierman19} and distributed control \cite{lin2022decentralized,bhuyan2024optimal} to name a few.

Traditionally, most works have considered the adversarial setting, providing worst-case guarantees against a dynamic hindsight optimal. Initial works \cite{BansalGupta15,Antoniadis18} provide a tight characterization for convex hitting costs in the one-dimensional setting. Higher dimensional analysis \cite{Antoniadis18,GoelWierman19} proves hardness of the adversarial setting in the absence of additional assumptions on structure of the hitting cost. Different aspects of the higher dimensional setting have been explored with significant advances in: (i) future predictions \cite{ChenWierman16,lin2022decentralized}, (ii) untrusted advice \cite{ChristiansonWierman23} and (iii) multi-agent scenarios \cite{lin2022decentralized,bhuyan2024optimal}.

\paragraph{\textit{Key Distinction}} A critical and near-universal assumption in these three areas is that the online algorithm has full knowledge of the current hitting cost function (or convex body) at each step \cite{GoelLinWierman19,GoelWierman19,ChristiansonWierman23,BhuyanMukherjee24}. While a few recent works have considered weaker feedback models like gradients \cite{senapati2023online} or delayed feedback \cite{senapati2023online,shah2025online}, they still assume a noiseless oracle. Our work departs significantly by addressing the SOCO problem under the much weaker and noisy zeroth-order (bandit) feedback model, where the cost function's structure must be learned online.

\subsection{Learning in Control: System Identification and LQG}

This area focuses on controlling a system while simultaneously learning its unknown dynamics \cite{Csaba11,9483309,10590739}. The central goal is to learn the parameters of a system's state-space model in an online fashion from input-output data, while incurring a known quadratic cost $x_t^T Q x_t + u_t^T R u_t$, over the taken actions $\{u_t\}_t$ and resulting state $\{x_{t+1}\}_t$. Recent work \cite{Lale2020,pmlr-v144-lale21b} has increasingly employed techniques from online learning to provide performance guarantees for unknown transition matrices $A$ and $B$ in a linear dynamical system $x_{t+1} = Ax_t+Bu_t+w_t$, where $w_t$ is an IID Gaussian drift. . This is crucial for applications in adaptive control \cite{Lale2020,Cohen2019}. Such a system, paired with the aforementioned quadratic costs, form the ``Linear Quadratic Gaussian'' (LQG) control paradigm. In the LQG setting, the goal is to control a linear system with quadratic costs under Gaussian drift. Regret minimization guarantees have been established with first-order feedback \cite{9483309,10590739}.

\paragraph{\textit{Key Distinction}} While these fields integrate learning and control, their primary focus is on learning the \textbf{system dynamics} (e.g., the state transition matrices), whereas our work focuses on learning the \textbf{cost function} itself. Furthermore, their theoretical guarantees often rely on assumptions that make unbounded movement a secondary concern, such as system controllability/stability \cite{Lale2020,9483309,10590739} or low-drift, sub-gaussian processes \cite{9483309,10590739}, which are assumptions we do not make.

\subsection{Online Learning and Online Convex Optimization (OCO)}
Bandit Convex Optimization (BCO) represents the extreme of limited information, where an algorithm must choose an action and only receives the resulting cost value (zeroth-order feedback) \cite{cesa2006price,lattimore2020bandit}. This aligns with our feedback model. Significant progress has been made in designing algorithms with optimal regret for this setting \cite{hazan2016optimal,liu2025non}. Direct application has been found in non-stochastic control, where adversarial disturbances are handled \cite{cassel2020bandit,sun2024tight}, albeit under a controllability assumption. A large body of work in OCO and BCO, provides sub-linear regret guarantees under two fundamental assumptions: (i) that the action set is a compact, bounded domain \cite{hazan2016optimal,cassel2020bandit,zhao2023non} and, (ii) a static benchmark for regret \cite{cassel2020bandit}. Only recently have works started addressing an adaptive benchmark \cite{zhao2023non} but still rely on bounded space and gradients.

\paragraph{\textit{Key Distinction}} The primary difference is the handling of switching costs. The vast majority of OCO and BCO literature does not include switching costs in the objective; the goal is simply to minimize the cumulative hitting cost. The assumption of a \textbf{bounded action space} common in this literature implicitly caps any potential movement, making explicit management of switching costs unnecessary. Our work addresses the more challenging problem of managing explicit, unbounded switching costs in an unbounded domain, a setting largely unexplored in the BCO literature.

\subsection{Inverse Optimal Control (IOC) and Inverse RL}
This research area is focused on the problem of inferring an agent's objectives (i.e., its cost or reward function) from demonstrations of its behavior. Inverse Optimal Control (IOC), particularly in the context of LQR, has seen significant progress in learning the underlying quadratic cost matrices ($Q$ and $R$) from observed state-action trajectories \cite{lian2023off}. This is highly relevant for applications in robotics \cite{priess2014solutions,el2019inverse} and human behavior modeling \cite{wu2023human,li5245076human}. The dominant paradigm in IOC and the broader Inverse Reinforcement Learning (IRL) field is offline and data-driven. State-of-the-art algorithms typically assume access to a pre-collected dataset of expert or hindsight-optimal trajectories from which to learn the cost function\cite{lian2023off,asl2025data}, where the formulation is ``off-policy'' or data-driven. Only very recently is there progress in on-policy IOC/IRL, where the dependence on expert trajectory is fractional (but not zero) \cite{sun2025inverse,liu2025intrajectory}.

\paragraph{\textit{Key Distinction}} The fundamental difference is the \textbf{online versus offline} nature of the problem. IOC/IRL learns from a static dataset of past behavior. Our algorithm operates \textbf{online}, where it must simultaneously learn the cost function and make decisions without the benefit of pre-existing expert demonstrations. The feedback our algorithm receives is generated by its own actions, creating a tight loop between exploration, estimation, and control that is absent in the offline IOC setting.

\section{Concluding Remarks}
In this work, we present \scale{} as the first algorithm that balances bandit exploration of dynamic quadratic hitting costs with movement-efficient exploitation to reduce unbounded metric switching costs, achieving a sub-linear dynamic regret in both rank-deficient and full-rank settings. Our key finding, the $\mathcal{O}(T^{2/3})$ regret for rank-deficient costs is the first in this space, made possible through a spectral perspective of traditional regret analysis. Future directions include regret guarantees for our hybrid exploitation scheme \hyscale{} in light tail environments and, extending the explore-then-exploit approach to adversarial settings.

\bibliographystyle{unsrtnat}
\bibliography{references}

\clearpage
\appendix
\thispagestyle{empty}

\section{Additional Numerical Experiments}\label{appendix:addn_exps}
We extensively test the performance of our proposed algorithm \scale{} and \hyscale{} against the three baselines: (i) Follow the Minimizer (\ftm{}), (ii) Passive Online Learner (\pol{}) and, (iii) Oracle Assisted Learner (\oal{}). In this section, we present all our findings.  
\subsection{Uncorrelated light-tailed process: Minimizer differences $\delta_t = v_{t-1}-v_t$ are IID Gausssian}
\begin{figure}[H]
    \centering
     \subfigure{%
        \includegraphics[width=0.32\linewidth]{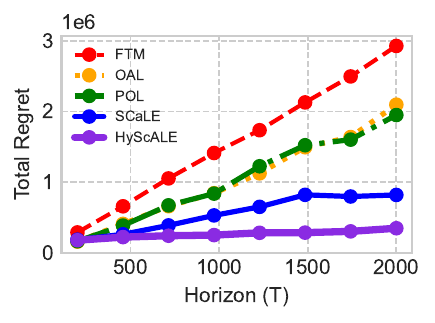}
        }
     \subfigure{%
    \includegraphics[width=0.32\linewidth]{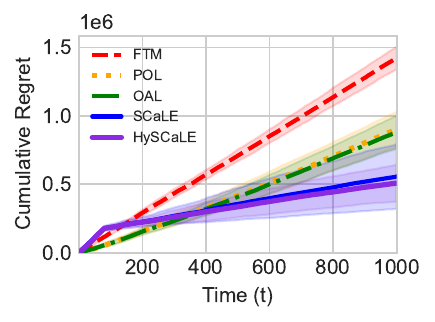}
    }
     \subfigure{%
\includegraphics[width=0.32\linewidth]{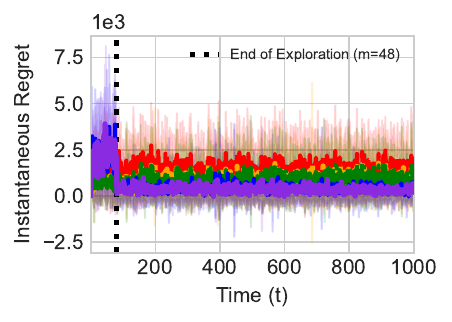}
    }
    \caption{$\regret_T$, $\regret_{1000}(t)$ and $\frac{\Delta \regret_{1000}(t)}{\Delta t}$, for $r=d=4, \sigma_r^A = 1, \sigma_v = 50,\bar{\eta} = 10$ and $c_1$ set to $3$}
    \label{fig:app_exp1_1}
    \vspace{-20pt}
\end{figure}

\begin{figure}[H]
    \centering
     \subfigure{%
        \includegraphics[width=0.32\linewidth]{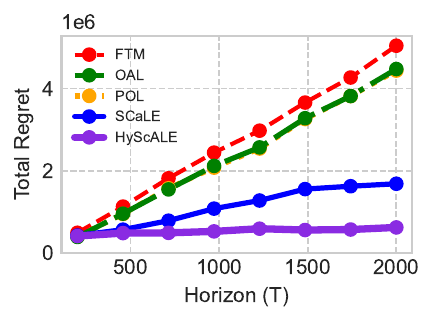}
        }
     \subfigure{%
    \includegraphics[width=0.32\linewidth]{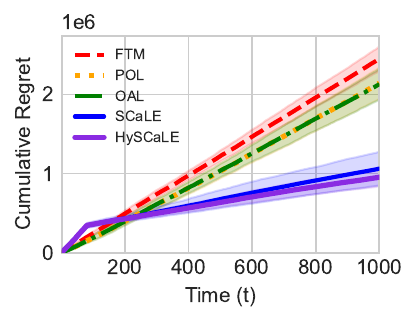}
    }
     \subfigure{%
\includegraphics[width=0.32\linewidth]{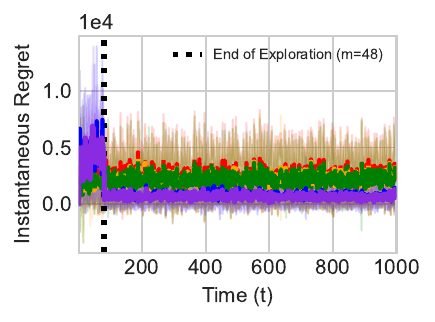}
    }
    \caption{$\regret_T$, $\regret_{1000}(t)$ and $\frac{\Delta \regret_{1000}(t)}{\Delta t}$, for $r=d=4, \sigma_r^A = 10^{-2}, \sigma_v = 50,\bar{\eta} = 10$ and $c_1$ set to $3$}
    \label{fig:app_exp_2}
    \vspace{-20pt}
\end{figure}

\begin{figure}[H]
    \centering
     \subfigure{%
        \includegraphics[width=0.32\linewidth]{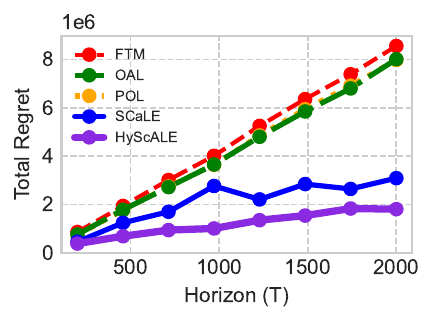}
        }
     \subfigure{%
    \includegraphics[width=0.32\linewidth]{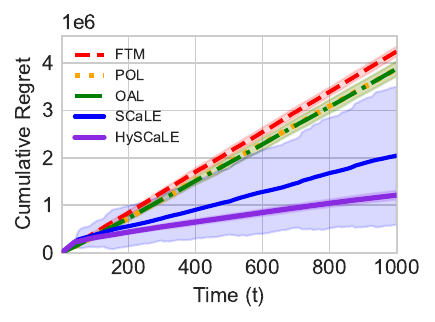}
    }
     \subfigure{%
\includegraphics[width=0.32\linewidth]{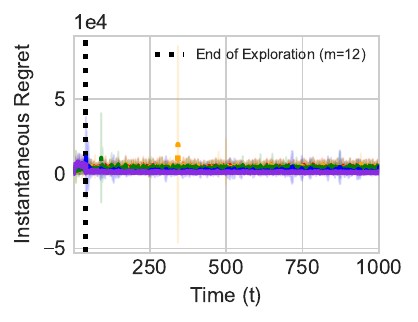}
    }
    \caption{$\regret_T$, $\regret_{1000}(t)$ and $\frac{\Delta \regret_{1000}(t)}{\Delta t}$, for $r=1, d=4, \sigma_r^A = 1, \sigma_v = 50,\bar{\eta} = 10$ and $c_1$ set to $3$}
    \label{fig:app_exp_3}
\end{figure}

\begin{figure}[H]
    \centering
     \subfigure{%
        \includegraphics[width=0.32\linewidth]{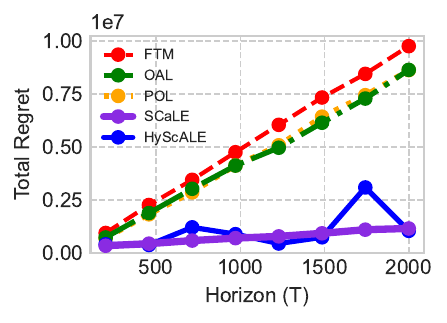}
        }
     \subfigure{%
    \includegraphics[width=0.32\linewidth]{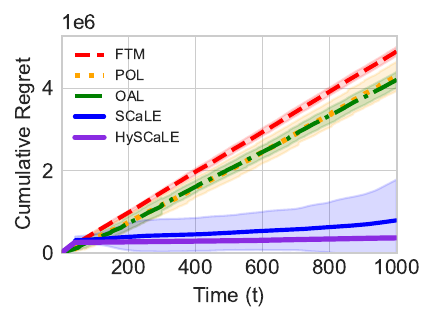}
    }
     \subfigure{%
\includegraphics[width=0.32\linewidth]{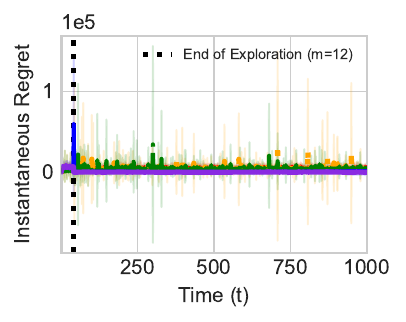}
    }
    \caption{$\regret_T$, $\regret_{1000}(t)$ and $\frac{\Delta \regret_{1000}(t)}{\Delta t}$, for $r=1, d=4, \sigma_r^A = 10^{-2}, \sigma_v = 50,\bar{\eta} = 10$ and $c_1$ set to $3$}
    \label{fig:app_exp_4}
\end{figure}

\subsection{Correlated light-tailed process: Non-IID increments $\delta_t = 0.3 \delta_{t-1}+\mathcal{N}(\mathbf{0},\sigma_v^2 I)$}

\begin{figure}[H]
    \centering
     \subfigure{%
        \includegraphics[width=0.32\linewidth]{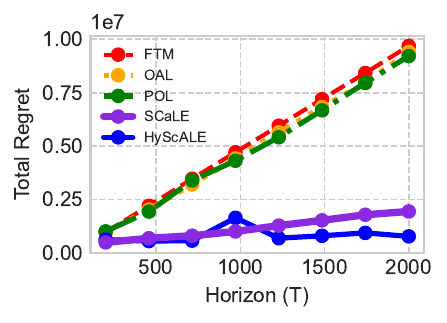}
        }
     \subfigure{%
    \includegraphics[width=0.32\linewidth]{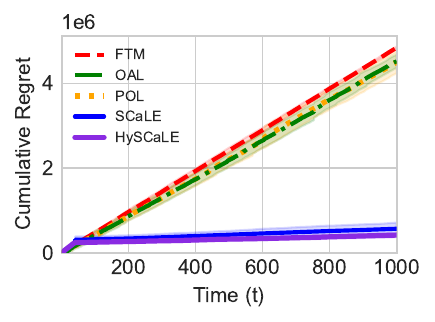}
    }
     \subfigure{%
\includegraphics[width=0.32\linewidth]{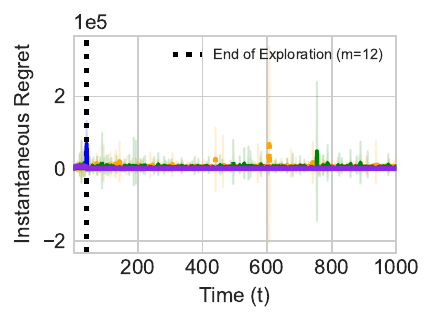}
    }
    \vspace{-10pt}
    \caption{$\regret_T$, $\regret_{1000}(t)$ and $\frac{\Delta \regret_{1000}(t)}{\Delta t}$, for $r=1,d=4, \sigma_r^A = 10^{-2}, \sigma_v = 50,\bar{\eta} = 10$ and $c_1$ set to $3$}
    \label{fig:app_exp_13}
\end{figure}

\begin{figure}[H]
    \centering
     \subfigure{%
        \includegraphics[width=0.32\linewidth]{figures/appendix/correlated/horizon_regret_linear_d4_r1_v50_c115.0_runs5_sigma0.01_seed0.pdf}
        }
     \subfigure{%
    \includegraphics[width=0.32\linewidth]{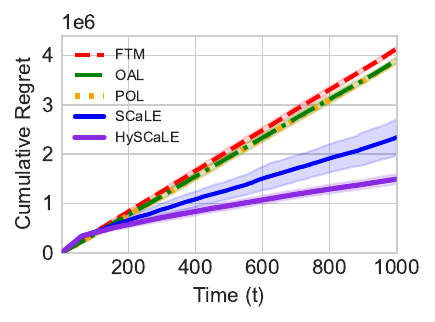}
    }
     \subfigure{%
\includegraphics[width=0.32\linewidth]{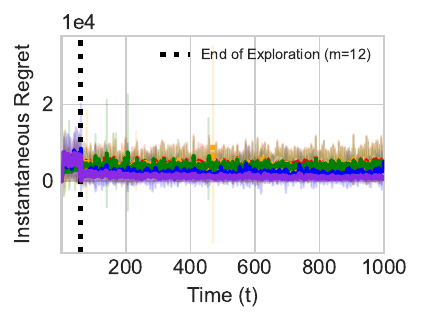}
    }
    \vspace{-10pt}
    \caption{$\regret_T$, $\regret_{1000}(t)$ and $\frac{\Delta \regret_{1000}(t)}{\Delta t}$, for $r=1, d=4, \sigma_r^A = 1, \sigma_v = 50,\bar{\eta} = 10$ and $c_1$ set to $3$}
    \label{fig:app_exp_14}
\end{figure}

\begin{figure}[H]
    \centering
     \subfigure{%
        \includegraphics[width=0.32\linewidth]{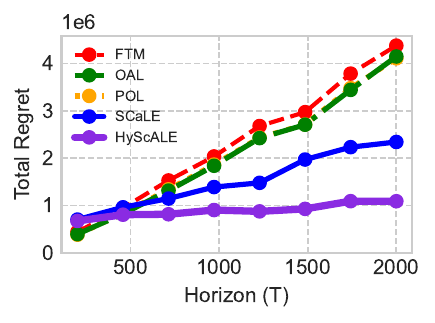}
        }
     \subfigure{%
    \includegraphics[width=0.32\linewidth]{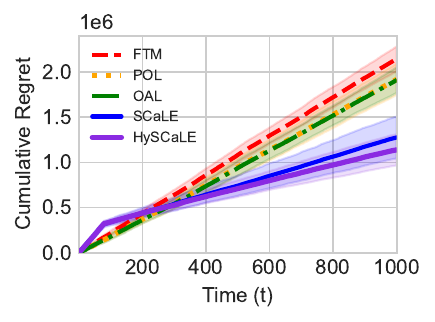}
    }
     \subfigure{%
\includegraphics[width=0.32\linewidth]{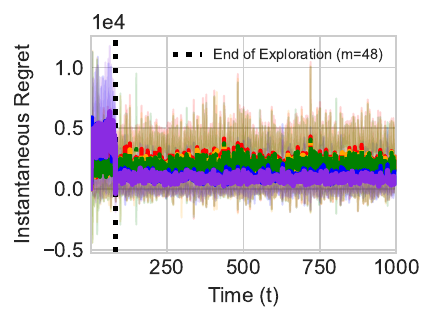}
    }
    \vspace{-10pt}
    \caption{$\regret_T$, $\regret_{1000}(t)$ and $\frac{\Delta \regret_{1000}(t)}{\Delta t}$, for $r=d=4, \sigma_r^A = 10^{-2}, \sigma_v = 50,\bar{\eta} = 10$ and $c_1$ set to $3$}
    \label{fig:app_exp_15}
    \vspace{-20pt}
\end{figure}

\begin{figure}[H]
    \centering
     \subfigure{%
        \includegraphics[width=0.32\linewidth]{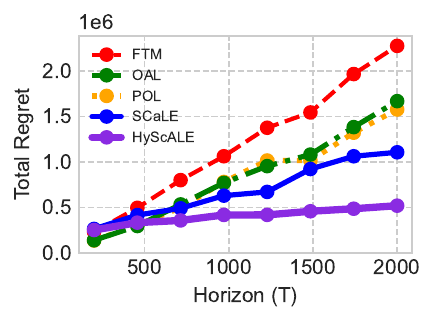}
        }
     \subfigure{%
    \includegraphics[width=0.32\linewidth]{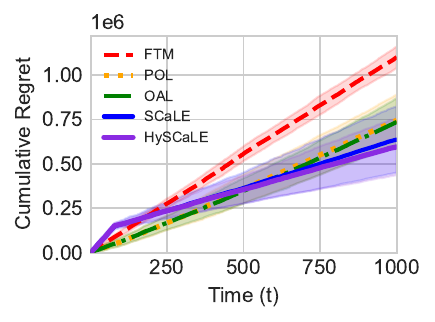}
    }
     \subfigure{%
\includegraphics[width=0.32\linewidth]{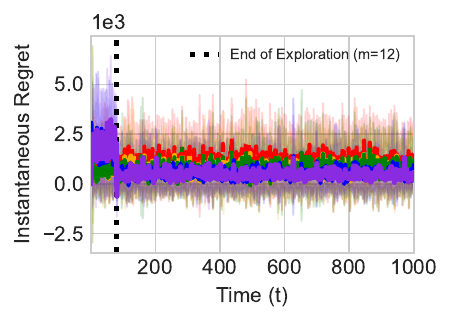}
    }
    \vspace{-10pt}
    \caption{$\regret_T$, $\regret_{1000}(t)$ and $\frac{\Delta \regret_{1000}(t)}{\Delta t}$, for $r=d=4, \sigma_r^A = 1, \sigma_v = 50,\bar{\eta} = 10$ and $c_1$ set to $3$}
    \label{fig:app_exp_16}
    \vspace{-10pt}
\end{figure}

\subsection{Fat-tails: Cauchy Distribution}
\vspace{-25pt}
\begin{figure}[H]
    \centering
     \subfigure{%
        \includegraphics[width=0.32\linewidth]{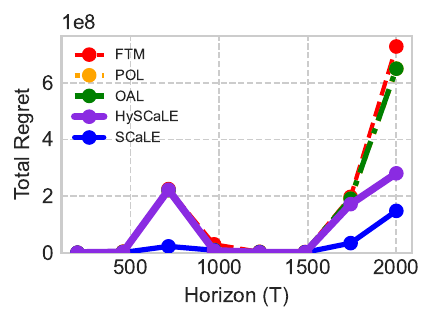}}
     \subfigure{%
    \includegraphics[width=0.32\linewidth]{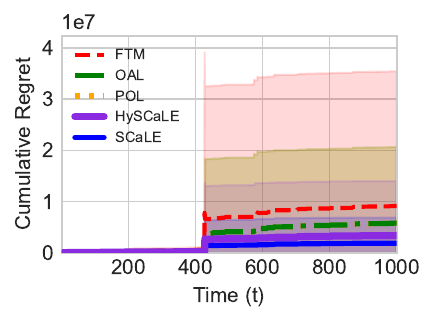}
    }
     \subfigure{%
\includegraphics[width=0.32\linewidth]{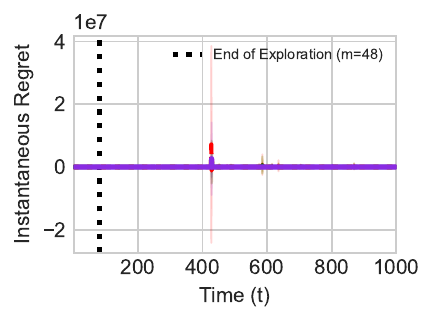}
    }
    \caption{$\regret_T$, $\regret_{1000}(t)$ and $\frac{\Delta \regret_{1000}(t)}{\Delta t}$, for $r=4, d=4, \sigma_r^A = 1, \sigma_v = 1,\bar{\eta} = 1$ and $c_1$ set to $5$}
    \label{fig:app_exp_5}
        \vspace{-20pt}
\end{figure}
\begin{figure}[H]
    \centering
     \subfigure{%
        \includegraphics[width=0.32\linewidth]{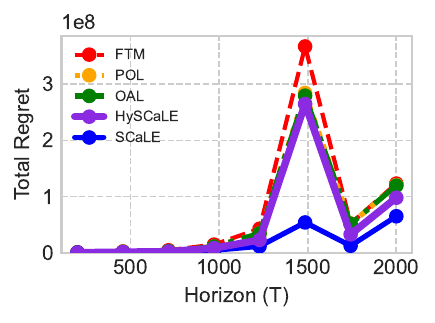}}
     \subfigure{%
    \includegraphics[width=0.32\linewidth]{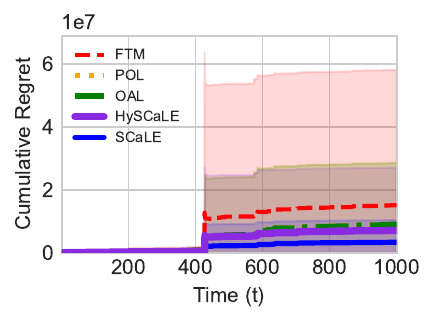}
    }
     \subfigure{%
\includegraphics[width=0.32\linewidth]{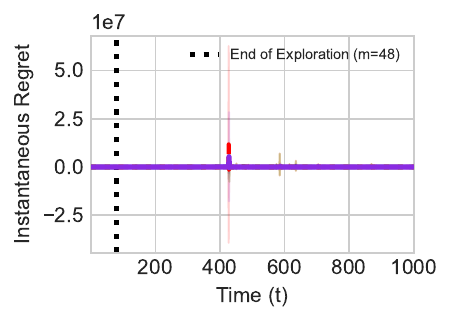}
    }
    \caption{$\regret_T$, $\regret_{1000}(t)$ and $\frac{\Delta \regret_{1000}(t)}{\Delta t}$, for $r=4, d=4, \sigma_r^A = 10^{-2}, \sigma_v = 1,\bar{\eta} = 1$ and $c_1$ set to $5$}
    \label{fig:app_exp_6}
\end{figure}
\begin{figure}[H]
    \centering
     \subfigure{%
        \includegraphics[width=0.32\linewidth]{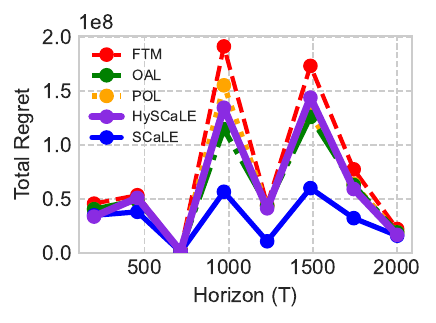}}
     \subfigure{%
    \includegraphics[width=0.32\linewidth]{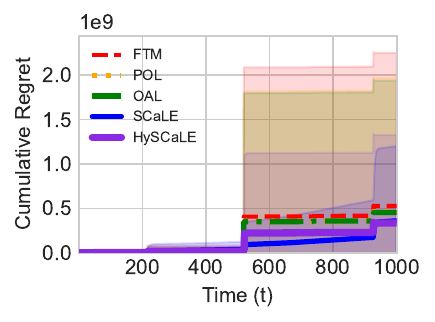}
    }
     \subfigure{%
\includegraphics[width=0.32\linewidth]{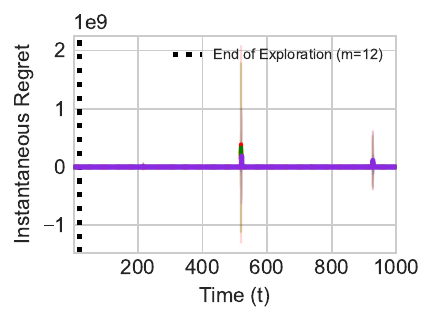}
    }
    \caption{$\regret_T$, $\regret_{1000}(t)$ and $\frac{\Delta \regret_{1000}(t)}{\Delta t}$, for $r=1, d=4, \sigma_r^A = 1, \sigma_v = 1,\bar{\eta} = 1$ and $c_1$ set to $5$}
    \label{fig:app_exp_7}
\end{figure}

\begin{figure}[H]
    \centering
     \subfigure{%
        \includegraphics[width=0.32\linewidth]{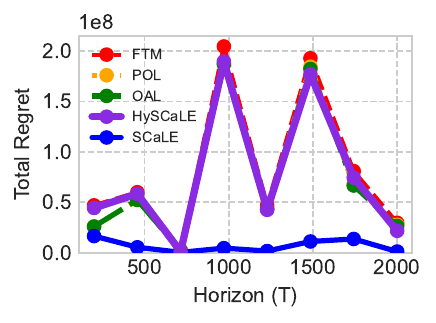}}
     \subfigure{%
    \includegraphics[width=0.32\linewidth]{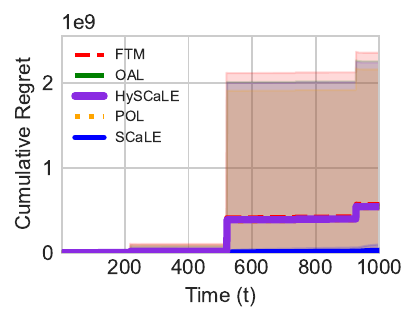}
    }
     \subfigure{%
\includegraphics[width=0.32\linewidth]{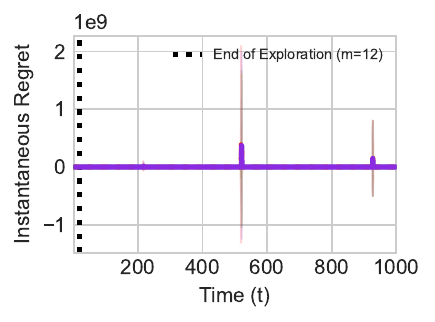}
    }
    \caption{$\regret_T$, $\regret_{1000}(t)$ and $\frac{\Delta \regret_{1000}(t)}{\Delta t}$, for $r=1, d=4, \sigma_r^A = 10^{-2}, \sigma_v = 1,\bar{\eta} = 1$ and $c_1$ set to $5$}
    \label{fig:app_exp_8}
\end{figure}

\subsection{Heavy-tails: Laplace Distribution}\label{appendix: laplace_exps}
\begin{figure}[H]
    \centering
     \subfigure{%
        \includegraphics[width=0.32\linewidth]{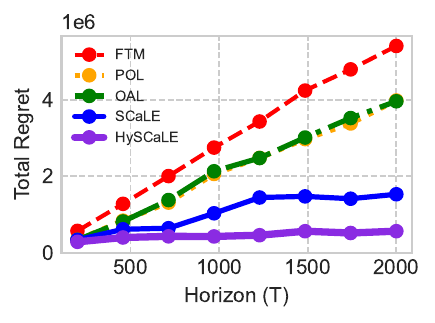}}
     \subfigure{%
    \includegraphics[width=0.32\linewidth]{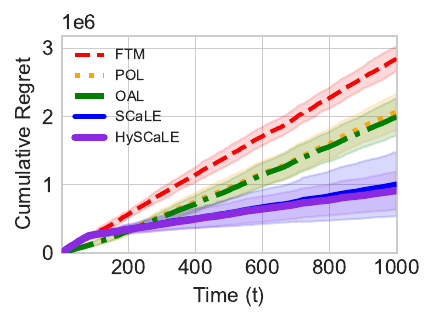}
    }
     \subfigure{%
\includegraphics[width=0.32\linewidth]{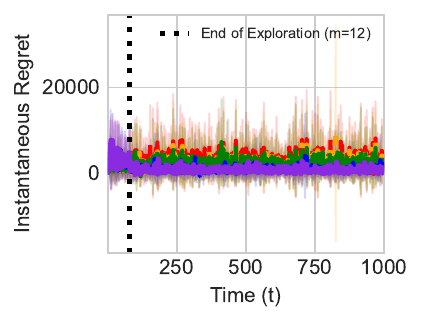}
    }
    \caption{$\regret_T$, $\regret_{1000}(t)$ and $\frac{\Delta \regret_{1000}(t)}{\Delta t}$, for $r=4, d=4, \sigma_r^A = 1, \sigma_v = 50,\bar{\eta} = 1$ and $c_1$ set to $5$}
    \label{fig:app_exp_9}
\end{figure}

\begin{figure}[H]
    \centering
     \subfigure{%
        \includegraphics[width=0.32\linewidth]{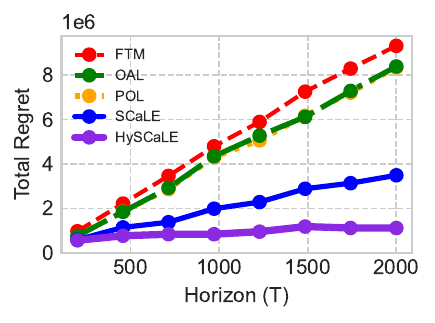}}
     \subfigure{%
    \includegraphics[width=0.32\linewidth]{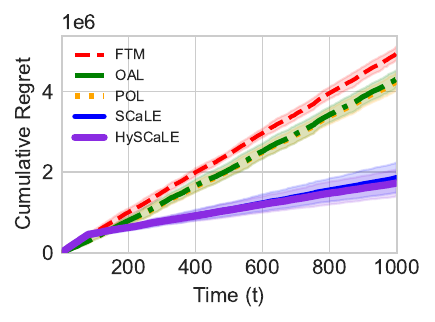}
    }
     \subfigure{%
\includegraphics[width=0.32\linewidth]{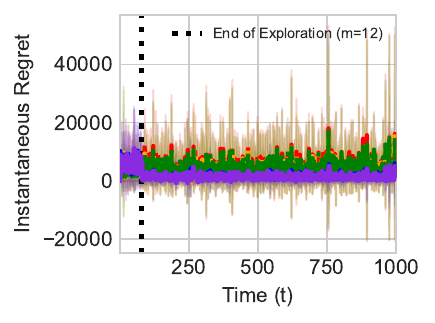}
    }
    \caption{$\regret_T$, $\regret_{1000}(t)$ and $\frac{\Delta \regret_{1000}(t)}{\Delta t}$, for $r=4, d=4, \sigma_r^A = 10^{-2}, \sigma_v = 50,\bar{\eta} = 1$ and $c_1$ set to $5$}
    \label{fig:app_exp_10}
\end{figure}
\begin{figure}[H]
    \centering
     \subfigure{%
        \includegraphics[width=0.32\linewidth]{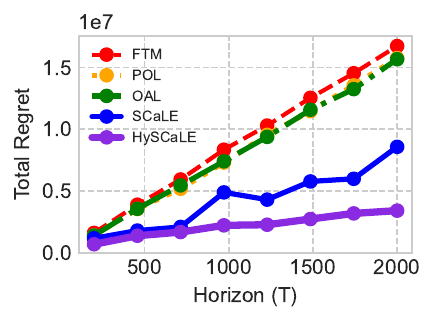}}
     \subfigure{%
    \includegraphics[width=0.32\linewidth]{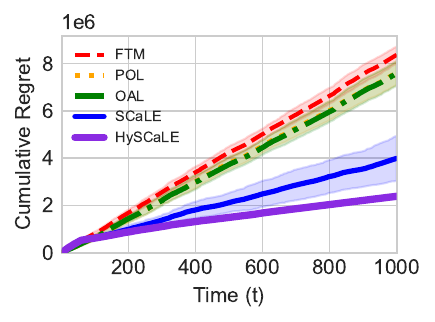}
    }
     \subfigure{%
\includegraphics[width=0.32\linewidth]{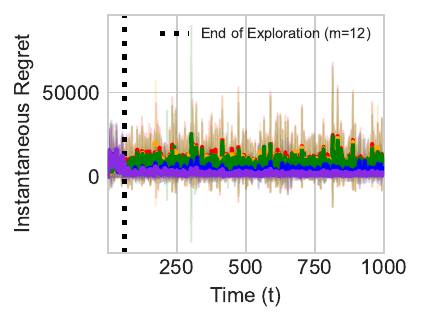}
    }
    \caption{$\regret_T$, $\regret_{1000}(t)$ and $\frac{\Delta \regret_{1000}(t)}{\Delta t}$, for $r=1, d=4, \sigma_r^A = 1, \sigma_v = 50,\bar{\eta} = 1$ and $c_1$ set to $5$}
    \label{fig:app_exp_11}
\end{figure}
\begin{figure}[H]
    \centering
     \subfigure{%
        \includegraphics[width=0.32\linewidth]{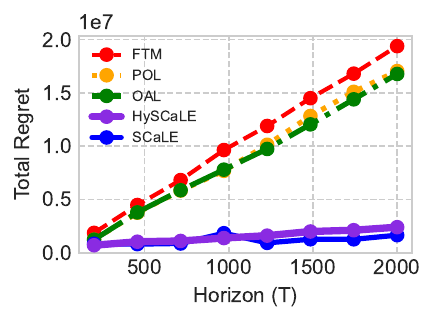}}
     \subfigure{%
    \includegraphics[width=0.32\linewidth]{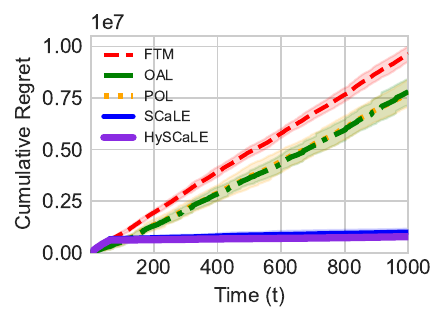}
    }
     \subfigure{%
\includegraphics[width=0.32\linewidth]{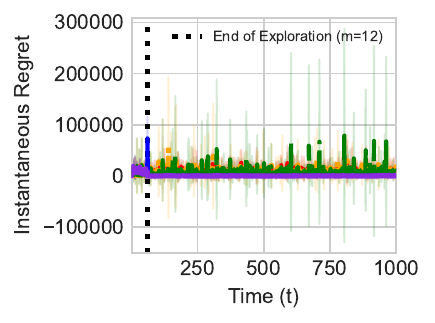}
    }
    \caption{$\regret_T$, $\regret_{1000}(t)$ and $\frac{\Delta \regret_{1000}(t)}{\Delta t}$, for $r=1, d=4, \sigma_r^A = 10^{-2}, \sigma_v = 50,\bar{\eta} = 1$ and $c_1$ set to $5$}
    \label{fig:app_exp_12}
\end{figure}

\section{Matrix Perturbation Basics}\label{appendix:matrix_perturb_basics}
Consider any $A = QDQ^T$ and its estimate $\hat{A} = P\hat{D}P^T.$ Both $A$ and $\hat{A}$ are positive semi-definite. WLOG $\lambda^{1,A}\geq \ldots \lambda^{d,A}$ and $\hat\lambda^{1}\geq \ldots \hat\lambda^{d}.$ We assume that $\|A-\hat{A}\|_*\leq \epsilon.$ Weyl's inequality for singular values establishes that 
\begin{align}
    |\hat{\lambda}^i-\lambda^{i,A}|\leq \|\hat{A}-A\|_{op} \leq \|\hat{A}-A\|_* \leq \epsilon 
\end{align}
which provides us with
\begin{align}
    \hat{\lambda}^i \in [\lambda^{i,A}-\epsilon,\lambda^{i,A}+\epsilon]
\end{align}
for $i\in \{1,\ldots,r\}$ and 
\begin{align}
     \hat{\lambda}^j \in [0,\epsilon]
\end{align}
for $j>r$ as $\lambda^{j,A} = 0$ and $\hat{A} \succeq 0$ Next we use the fact that $\|\hat{A}-A\|_F \leq \|\hat{A}-A\|_*/\sqrt{d}$ to establish that
\begin{align}
    |\hat{A}_{i,j} - A_{i,j}| \leq \epsilon.
\end{align}
For $\hat{A} = A + Z$ (where $Z = \hat{A}-A$, we have from \cite{cai2018rate}
\begin{align}
    \frac{1}{\sqrt{2}}\|P-Q\|_{op} \leq \|\sin \Theta(P,Q)\|_{op} \leq \frac{\alpha z_{12}+\beta z_{21}}{\alpha^2 -\beta^2 - \min(z_{12}^2,z_{21}^2)}
\end{align}
where $z_{ij} \leq \|Z\|_{F}$, $\alpha = \sigma_{\min}^A - \epsilon$ and $\beta \leq \epsilon.$ This implies
\begin{align}
    \|P-Q\|_{op} \lessapprox\frac{\sqrt{2}\cdot \epsilon}{\sigma_{\min}^A -\epsilon} 
\end{align}

\section{Matrix Perturbation driven Regret Analysis Framework}
\subsection{Proof of Proposition \ref{prop:spectral_split}}\label{proof:spectral_split}
At round $T$, the cost of the approximate policy is
\begin{align}
    \text{Cost}[T,T] &= \frac{1}{2}\|x_T-v_T\|_A^2+\frac{1}{2}\|x_T-x_{T-1}\|_2^2 \\
    &= \frac{1}{2}\|x_{T-1}-v_T\|_{\hat{C}_T A \hat{C}_T}^2+\frac{1}{2}\|v_T-x_{T-1}\|_{(I-\hat{C}_T)^2}^2\\
    &= \frac{1}{2}\|x_{T-1}-v_T\|_{\hat{C}_T \hat{A} \hat{C}_T}^2+\frac{1}{2}\|x_T-x_{T-1}\|_{(I-\hat{C}_T)^2}^2 + \frac{1}{2}\|x_{T-1}-v_T\|_{\hat{C}_T(A-\hat{A})\hat{C}_T}^2\\
    &=  \frac{1}{2}\|x_{T-1}-v_T\|_{I-\hat{C}_T}^2 + \underbrace{\frac{1}{2}\|x_{T-1}-v_T\|_{\hat{C}_T(A-\hat{A})\hat{C}_T}^2}_{\Delta_T}
\end{align}
Taking expectation with respect to $\F_{T-1}$, we get
\begin{align}
    \E[\text{Cost}[T,T] | \F_{T-1}] \leq \frac{1}{2}\|x_{T-1}-v_{T-1}\|_{I-\hat{C}_T}^2 + \frac{1}{2}\E\|v_{T-1}-v_{T}\|_{I-\hat{C}_T}^2 + \E[\Delta_T|\F_{T-1}].
\end{align}
We claim that
\begin{align}
    \E[\text{Cost}[T-t+1,T] | \F_{T-t}] \leq \frac{1}{2}\|x_{T-t}-v_{T-t}\|_{I-\hat{C}_{T-t+1}}^2 + \sum_{s=0}^{t-1} \frac{1}{2}\E\|v_{T-s-1}-v_{T-s}\|_{I-\hat{C}_{T-s}}^2 + \sum_{s=0}^{t-1} \E[\Delta_{T-s}|\F_{T-t}].
\end{align}
The above claim holds for $t=1.$ Suppose we it holds for some $t\geq 1$, then
\begin{align}
    &\frac{1}{2}\|x_{T-t}-v_{T-t}\|_A^2+\frac{1}{2}\|x_{T-t}-x_{T-t-1}\|_2^2 + \frac{1}{2}\|x_{T-t}-v_{T-t}\|_{I-\hat{C}_{T-t+1}}^2 \\
    &= \frac{1}{2}\|x_{T-t-1}-v_{T-t}\|_{\hat{C}_{T-t} (I+A-\hat{C}_{T-t}) \hat{C}_{T-t}}^2+\frac{1}{2}\|v_{T-t}-x_{T-t-1}\|_{(I-\hat{C}_{T-t})^2}^2\\
    &= \frac{1}{2}\|x_{T-t-1}-v_{T-t}\|_{\hat{C}_{T-t}(I+\hat{A}-\hat{C}_{T-t}) \hat{C}_{T-t}}^2+\frac{1}{2}\|v_{T-t}-x_{T-t-1}\|_{(I-\hat{C}_{T-t})^2}^2 + \frac{1}{2}\|x_{T-t-1}-v_{T-t}\|_{\hat{C}_{T-t}(A-\hat{A})\hat{C}_{T-t}}^2\\
    &=  \frac{1}{2}\|x_{T-t-1}-v_{T-t}\|_{I-\hat{C}_{T-t}}^2 + \underbrace{\frac{1}{2}\|x_{T-t-1}-v_{T-t}\|_{\hat{C}_{T-t}(A-\hat{A})\hat{C}_{T-t}}^2}_{\Delta_{T-t}}\\
\end{align}
which leads to
\begin{align}
    &\E[\text{Cost}[T-t,T] | \F_{T-t-1}]\\
    &= \E\left[\frac{1}{2}\|x_{T-t}-v_{T-t}\|_A^2+\frac{1}{2}\|x_{T-t}-x_{T-t-1}\|_2^2 \bigg|\F_{T-t-1}\right] + \E[\text{Cost}[T-t+1,T] | \F_{T-t-1}]\\
    &= \frac{1}{2}\|x_{T-t-1}-v_{T-t-1}\|_{I-\hat{C}_{T-t}}^2 + \sum_{s=0}^{t} \frac{1}{2}\E\|v_{T-s-1}-v_{T-s}\|_{I-\hat{C}_{T-s}}^2 + \sum_{s=0}^{t} \E[\Delta_{T-s}|\F_{T-t-1}]
\end{align}
proving our claim by induction. We, therefore, get an upper bound on the total cost as
\begin{align}
    \E[\text{Cost}[1,T] | \F_{0}] = \sum_{s=1}^{T} \frac{1}{2}\E\|v_{s}-v_{s-1}\|_{I-\hat{C}_{s}}^2 + \sum_{k=1}^{T} \E[\Delta_{k}|\F_{0}]
\end{align}
which gives a regret that is \textbf{\textit{equal to}}
\begin{align}
    \text{Regret}_T &= \sum_{s=1}^{T} \frac{1}{2}\E\|v_{s}-v_{s-1}\|_{C_{s}-\hat{C}_{s}}^2 +  \sum_{k=1}^{T} \frac{1}{2}\E\|x_{k-1}-v_k\|_{\hat{C}_k(A-\hat{A})\hat{C}_k}^2
\end{align}

\subsection{Proof of Theorem \ref{thm:exploit_regret_full}}\label{proof:exploit_regret_full}
The algorithm is
\begin{align}
    x_t = \Hat{C}_t x_{t-1}+(I-\hat{C}_t)v_t
\end{align}
where at round $t$, the player makes estimations
\begin{align}
    \hat{C}_{\tau}^{-1} = 2I+\hat{A} - \hat{C}_{\tau+1}
\end{align}
for $\tau\geq t.$ The gap to the DP sequence:
\begin{align}
    C_t^{-1} = 2I+A-C_{t+1}
\end{align}
can be quantified as
\begin{align}
    \hat{C}_{t} - C_t = \sum_{\tau=t}^T \prod_{m=t}^\tau C_{m} (A-\hat{A}) \prod_{m=\tau}^t \hat{C}_{m}
\end{align}
which for $\sigma^A_{\min}>0$ can be quantified as
\begin{align}
    \|\hat{C}_{t} - C_t\|_{op} \leq \frac{\|A-\hat{A}\|_{op}}{\sigma^A_{\min}} \leq \frac{\epsilon}{\sigma^A_{\min}}
\end{align}
Recall the regret is exactly equal to
\begin{align}
    \regret_T = \underbrace{\sum_{s=1}^{T} \frac{1}{2}\E\|v_{s}-v_{s-1}\|_{C_{s}-\hat{C}_{s}}^2}_{R_1} +  \underbrace{\sum_{s=1}^{T} \frac{1}{2}\E\|x_{s-1}-v_s\|_{\hat{C}_s(A-\hat{A})\hat{C}_s}^2}_{R_2}
\end{align}
For $R_1,$ we have
\begin{align}
    R_1 &\leq \sum_{s=1}^T\frac{\epsilon}{2\sigma^A_{\min}}\E\|v_s-v_{s-1}\|_2^2\\
    &\leq \frac{tr(\Sigma)}{2\sigma^A_{\min}}\cdot \epsilon T
\end{align}
For $R_2,$ we look at each term
\begin{align}
    \E\|x_{s-1}-v_{s}\|_{\hat{C}_{s} (A-\hat{A}) \hat{C}_{s}}^2 = \sum_{k=0}^{s-1} \E\|v_{s-k-1}-v_{s-k}\|_{\prod_{l=k}^0\hat{C}_{s-l} (A-\hat{A}) \prod_{l=0}^k\hat{C}_{s-l}}^2.
\end{align}
We look at
\begin{align}
    &\prod_{l=k}^0\hat{C}_{s-l} (A-\hat{A}) \prod_{l=0}^k\hat{C}_{s-l} \\
    =& P\times diag\left(\prod_{l=k}^0 \hat{\lambda}^1_{s-l},\ldots,\prod_{l=k}^0 \hat{\lambda}^d_{s-l}\right)\times \underbrace{(P^T Q D Q^T P - \hat{D})}_{P^T Q D Q^T P - D + D - \hat{D}} \times diag\left( \prod_{l=k}^0 \hat{\lambda}^1_{s-l},\ldots,\prod_{l=k}^0 \hat{\lambda}^d_{s-l}\right) \times P^T.\\
    =& \underbrace{P\times diag\left(\prod_{l=k}^0 \hat{\lambda}^1_{s-l},\ldots,\prod_{l=k}^0 \hat{\lambda}^d_{s-l}\right)\times \underbrace{(P^T Q D Q^T P - D)}_{\Delta_2} \times diag\left(\prod_{l=k}^0 \hat{\lambda}^1_{s-l},\ldots,\prod_{l=k}^0 \hat{\lambda}^d_{s-l}\right) P^T}_{X_1}\\
    &+  \underbrace{P\times diag\left(\prod_{l=k}^0 \hat{\lambda}^1_{s-l},\ldots,\prod_{l=k}^0 \hat{\lambda}^d_{s-l}\right)\times (D-\hat{D}) \times diag\left(\prod_{l=k}^0 \hat{\lambda}^1_{s-l},\ldots,\prod_{l=k}^0 \hat{\lambda}^d_{s-l}\right) P^T}_{X_2}.
\end{align}
Lets look at $X_2$ first:
\begin{align}
    D - \hat{D} &= diag(\lambda^{A,1}-\hat{\lambda}^1,\ldots,\lambda^{A,d}-\hat{\lambda}^d)\\
    &\preceq \epsilon I_{d\times d}
\end{align}
Also, the following holds for each $i$
\begin{align}
    \hat{\lambda}^i_t \leq \frac{1}{1+\lambda^{A,i}-\epsilon}
\end{align}
which will give us
\begin{align}
    &diag\left(\prod_{l=k}^0 \hat{\lambda}^1_{s-l},\ldots,\prod_{l=k}^0 \hat{\lambda}^d_{s-l}\right) \times (D - \hat{D}) \times diag\left(\prod_{l=k}^0 \hat{\lambda}^1_{s-l},\ldots,\prod_{l=k}^0 \hat{\lambda}^d_{s-l}\right)\\
    &\preceq \underbrace{diag\left(\frac{\epsilon}{(1+\lambda^{A,1}-\epsilon)^{2k+2}},\ldots,\frac{\epsilon}{(1+\lambda^{A,r}-\epsilon)^{2k+2}},0,\ldots,0\right)}_{\Delta_1}
\end{align}
leading to
\begin{align}
    \E\|v_{s-k-1}-v_{s-k}\|_{X_2}^2 &= \langle X_2,\Sigma\rangle\\
    &= \langle \Delta_1, P^T\Sigma P\rangle\\
    &= \sum_{i=1}^r \frac{\epsilon}{(1+\lambda^{A,i}-\epsilon)^{2k+2}} p_i^T \Sigma p_i\\
    &\leq \frac{\epsilon}{(1+\sigma^A_{\min}-\epsilon)^{2k+2}}tr(P^T\Sigma P)\\
    &= \epsilon \cdot \frac{tr(\Sigma)}{(1+\sigma^A_{\min}-\epsilon)^{2k+2}}
\end{align}
We will now look at $X_1$, and more specifically, $\Delta_2$:
\begin{align}
    \Delta_2 &= P^T Q D Q^T P - D\\
    &= (P^T Q-I + I) D (Q^T P -I +I) -D\\
    &= (P^T Q-I)D (Q^T P -I) + (P^T Q-I)D + D (Q^T P -I) + D-D\\
    &= \underbrace{(P^T Q-I)D (Q^T P -I)}_{\Delta_{2a}} + \underbrace{(P^T Q-I)D}_{\Delta_{2b}} + \underbrace{D (Q^T P -I)}_{\Delta_{2c}}
\end{align}
Now, observe that
\begin{align}
     \Delta_{3b} &= diag\left(\prod_{l=k}^0 \hat{\lambda}^1_{s-l},\ldots,\prod_{l=k}^0 \hat{\lambda}^d_{s-l}\right) \times \Delta_{2b}  \times diag\left(\prod_{l=k}^0 \hat{\lambda}^1_{s-l},\ldots,\prod_{l=k}^0 \hat{\lambda}^d_{s-l}\right)\\
     &= diag\left(\prod_{l=k}^0 \hat{\lambda}^1_{s-l},\ldots,\prod_{l=k}^0 \hat{\lambda}^d_{s-l}\right) \times (P^TQ-I) \times diag\left(\lambda^{A,1}\prod_{l=k}^0 \hat{\lambda}^1_{s-l},\ldots,\lambda^{A,d}\prod_{l=k}^0 \hat{\lambda}^d_{s-l}\right) 
\end{align}
and
\begin{align}
     \Delta_{3c} &=diag\left(\prod_{l=k}^0 \hat{\lambda}^1_{s-l},\ldots,\prod_{l=k}^0 \hat{\lambda}^d_{s-l}\right) \times \Delta_{2c}  \times diag\left(\prod_{l=k}^0 \hat{\lambda}^1_{s-l},\ldots,\prod_{l=k}^0 \hat{\lambda}^d_{s-l}\right)\\
     &= diag\left(\lambda^{A,1}\prod_{l=k}^0 \hat{\lambda}^1_{s-l},\ldots,\lambda^{A,d}\prod_{l=k}^0 \hat{\lambda}^d_{s-l}\right) \times (P^TQ-I)  diag\left(\prod_{l=k}^0 \hat{\lambda}^1_{s-l},\ldots,\prod_{l=k}^0 \hat{\lambda}^d_{s-l}\right).  
\end{align}
Lastly, $\Delta_{3a}$ is defined accordingly,
\begin{align}
    \Delta_{3a} &=diag\left(\prod_{l=k}^0 \hat{\lambda}^1_{s-l},\ldots,\prod_{l=k}^0 \hat{\lambda}^d_{s-l}\right) \times \Delta_{2a}  \times diag\left(\prod_{l=k}^0 \hat{\lambda}^1_{s-l},\ldots,\prod_{l=k}^0 \hat{\lambda}^d_{s-l}\right).
\end{align}
Putting these three together now
\begin{align}
     \E\|v_{s-k-1}-v_{s-k}\|_{X_1}^2 &= \langle \Delta_{3a},\P^T \Sigma P\rangle + \langle \Delta_{3b},\P^T \Sigma P\rangle + \langle \Delta_{3c},P^T \Sigma P\rangle\\
     &\leq \|\Sigma\|_F\cdot (\|\Delta_{3a}\|_F+\|\Delta_{3b}\|_F+\|\Delta_{3c}\|_F)
\end{align}
Now, $\|P^TQ-I\|_F = \|Q^TP-I\|_F = \|P-Q\|_F\leq \sqrt{d}\cdot\|P-Q\|_{op}\leq \frac{\sqrt{2d}\cdot \epsilon}{\sigma^A_{\min}-\epsilon}$. We, therefore, have
\begin{align}
    \|\Delta_{3b}\|_F = \|\Delta_{3c}\|_F \leq \frac{\sigma^A_{\max} \sqrt{2d} \cdot\epsilon}{(\sigma^A_{\min}-\epsilon)(1+\sigma^A_{\min}-\epsilon)^{2k+2}}
\end{align}
and
\begin{align}
    \|\Delta_{3a}\|_F &\leq \frac{1}{(1+\sigma^A_{\min}-\epsilon)^{2k+2}} \|\Delta_{2a}\|_F\\
    &\leq \frac{1}{(1+\sigma^A_{\min}-\epsilon)^{2k+2}} \|P^TQ-I\|_F \cdot \|D(Q^TP-I)\|_F\\
    &\leq \frac{1}{(1+\sigma^A_{\min}-\epsilon)^{2k+2}} \frac{2d\sigma^A_{\max} \cdot \epsilon^2}{(\sigma^A_{\min}-\epsilon)^2}.
\end{align}
Now putting $X_1$ and $X_2$ together,
\begin{align}
    &\E\|x_{s-1}-v_{s}\|_{\hat{C}_{s} (A-\hat{A}) \hat{C}_{s}}^2 \\
    &\leq \sum_{k=0}^{s-1} \frac{2d\sigma^A_{\max} \|\Sigma\|_F \cdot \epsilon^2}{(\sigma^A_{\min}-\epsilon)^2 (1+\sigma^A_{\min}-\epsilon)^{2k+2}} + 2\sum_{k=0}^{s-1} \frac{\|\Sigma\|_F\sigma^A_{\max} \sqrt{2d} \cdot\epsilon}{(\sigma^A_{\min}-\epsilon)(1+\sigma^A_{\min}-\epsilon)^{2k+2}} +\sum_{k=0}^{s-1} \epsilon \cdot \frac{tr(\Sigma)}{(1+\sigma^A_{\min}-\epsilon)^{2k+2}}\\
    &\leq \frac{1}{(1+\sigma^A_{\min}-\epsilon)^2-1}\left(\frac{2d\sigma^A_{\max} \|\Sigma\|_F}{(\sigma^A_{\min}-\epsilon)^2}\cdot\epsilon^2+\frac{2\|\Sigma\|_F\sigma^A_{\max} \sqrt{2d} \cdot\epsilon}{\sigma^A_{\min}-\epsilon}+\frac{tr(\Sigma) \epsilon}{(1+\sigma^A_{\min}-\epsilon)^2-1}\right)
\end{align}
which leads to a regret of 
\begin{align}
\begin{split}
    \text{Regret}[1,T] \leq& \frac{\epsilon T}{(1+\sigma^A_{\min}-\epsilon)^2-1}\left(\frac{2d\sigma^A_{\max} \|\Sigma\|_F}{(\sigma^A_{\min}-\epsilon)^2}\cdot\epsilon+\frac{2\|\Sigma\|_F\sigma^A_{\max} \sqrt{2d}}{\sigma^A_{\min}-\epsilon}+\frac{tr(\Sigma)}{(1+\sigma^A_{\min}-\epsilon)^2-1}\right) \\
    &+ \frac{tr(\Sigma)}{2\sigma^A_{\min}}\cdot \epsilon T.
\end{split}
\end{align}
\subsection{Proof of Theorem \ref{thm:regret_decomp}}\label{proof:regret_decomp}
Now, if the first $m$ rounds involve data collection through the action sequence:
\begin{align}
    x_t = v_t + \gamma\cdot z_t
\end{align}
with $z_t \sim \mathcal{N}(\mathbf{0},I),$ the cost involved in the process is
\begin{align}
    \sum_{t=1}^m \frac{1}{2}&\E\|x_t-v_t\|_A^2 + \frac{1}{2}\E\|x_t-x_{t-1}\|_2^2\\
    &= \sum_{t=1}^m\frac{\gamma^2}{2}\E[z_t^T A z_t] + \frac{1}{2}\E\|v_t+\gamma z_t - v_{t-1} -\gamma z_{t-1}\|_2^2\\
    &= \sum_{t=1}^m\frac{\gamma^2}{2}\E\langle A,z_t z_t^T\rangle + \frac{1}{2}\E\|v_t - v_{t-1}\|_2^2 + \frac{\gamma^2}{2}\E\|z_t\|_2^2 + \frac{\gamma^2}{2}\E\|z_{t-1}\|_2^2\\
    &= \sum_{t=1}^m\frac{\gamma^2}{2}\langle A,I\rangle + (1/2)tr(\Sigma) + \gamma^2 d\\
    &= m\cdot(1/2)tr(\Sigma) + m\gamma^2 \cdot \left((1/2)tr(A)+d\right)
\end{align}
These measurements are used by the player in the form:
\begin{align}
    y_t &= \frac{\gamma^2}{2} z_t^T A z_t + \eta_t\\
    \frac{y_t}{\gamma^2} &= \frac{1}{2}z_t^T A z_t + \frac{\eta_t}{\gamma^2}\\
    y_t' &= \frac{1}{2}\langle A,z_t z_t^T\rangle + \eta_t'
\end{align}
where the total measurement error sums up to $\epsilon_m = \sum_{t=1}^m |\eta_t'| = \frac{\|\bm{\eta}_{1:m}\|_1}{\gamma^2}$. Solving for
$$
    \hat{A} = \argmin_{\substack{M \succeq 0\\ \|Y'-\mathcal{A}(M)\|_1 \leq \epsilon_m}} \quad Tr(M)
$$
we are guaranteed that $\|\hat{A}-A\|_* \leq \frac{\epsilon_m}{m} = \frac{\|\bm{\eta}_{1:m}\|_1}{m\gamma^2}$ with probability $1-\exp(-C_0m)$ where $C_0$ is a universal constant. This holds only for $m>c_1\cdot d^2$ for a known universal constant $c_1.$ The final regret guarantee we get using this is:
\begin{align}
    \text{Regret}[1,T]=& \regret[1,m] + \frac{1}{2}\|x_{m}-v_{m}\|_{I-\hat{C}_{m+1}}^2 + \sum_{s=m+1}^{T} \frac{1}{2}\E\|v_{s}-v_{s-1}\|_{I-\hat{C}_{s}}^2 + \frac{1}{2}\E\|x_{s-1}-v_s\|_{\hat{C}_s(A-\hat{A})\hat{C}_s}^2\\
    =& \regret[1,m] + \frac{1}{2}\|x_{m}-v_{m}\|_{I-\hat{C}_{m+1}}^2 + \regret[m+1,T]\\
    \begin{split}
    \leq& m\gamma^2 \cdot \left((1/2)tr(A)+d\right)+ \sum_{t=1}^m \left\{(1/2)\underbrace{tr(\Sigma)}_{\langle \Sigma,I\rangle} -\frac{1}{2}\underbrace{\E\|v_{s-1}-v_{s}\|_{I-C_s}^2}_{\langle\Sigma,I-C_s\rangle} \right\} + \frac{\gamma^2}{2}d\\
    &+\regret[m+1,T]
    \end{split}\\
    \begin{split}
    =& m\gamma^2 \cdot \left((1/2)tr(A)+d\right) + \sum_{t=1}^m (1/2)\langle \Sigma,C_s\rangle  + \frac{\gamma^2}{2}d + \regret[m+1,T]
    \end{split}\\
    \begin{split}
    \leq& \underbrace{m\left(\frac{(1/2)tr(\Sigma)}{1+\sigma^A_{\min}}\right)}_{\text{Irreducible Cost of NOT knowing } A} + \underbrace{m\gamma^2 \cdot \left((1/2)tr(A)+d\right)+ \frac{\gamma^2}{2}d}_{\text{Additional cost of data collection}}  + \underbrace{\regret[m+1,T]}_{o(T) \text{ Regret}}
    \end{split}
\end{align}
\subsection{Proof of Corollary \ref{corr:equal-contribution}}\label{proof:equal-contribution}
For the full-rank case, that is $rank(A) = r=d$, we have
\begin{align}
\begin{split}
    \text{Regret}[m+1,T] \leq& \frac{\epsilon (T-m)}{(1+\sigma^A_{\min}-\epsilon)^2-1}\left(\frac{2d\sigma^A_{\max} \|\Sigma\|_F}{(\sigma^A_{\min}-\epsilon)^2}\cdot\epsilon+\frac{2\|\Sigma\|_F\sigma^A_{\max} \sqrt{2d}}{\sigma^A_{\min}-\epsilon}+\frac{tr(\Sigma)}{(1+\sigma^A_{\min}-\epsilon)^2-1}\right) \\
    &+ \frac{tr(\Sigma)}{2\sigma^A_{\min}}\cdot \epsilon (T-m).
\end{split}
\end{align}
 where $\epsilon = \|\hat{A}-A\|_* \leq \frac{\|\bm{\eta}_{1:m}\|_1}{m\gamma^2}$. Considering the worst noise model, where total noise scales linearly with the measurements: $\|\bm{\eta}_{1:m}\|_1 = \bar{\eta}m,$ we choose $\gamma^2 = \sqrt{\bar{\eta}}\max\left\{(T-m)^{1/2},\frac{2}{\sigma^A_{\min}}\right\} \implies \epsilon \leq \min\left\{\frac{1}{\sqrt{T-m}},\frac{\sigma^A_{\min}}{2}\right\}$ and $m$ as the minimum number of rounds for consistency of the estimate $\hat{A}$, that is, $m = c_1\cdot d^2$. This gives a regret guarantee
 \begin{align}
     \begin{split}
    \regret_T \leq& \sqrt{\bar{\eta}(T-c_1d^2)} \left( \frac{4\sigma^A_{\max} \|\Sigma\|_F (d+\sqrt{2d}) + tr(\Sigma)}{(\sigma^A_{\min})^2} + \frac{tr(\Sigma)}{2\sigma^A_{\min}} \right) \\
    &+ \max\left\{\sqrt{\bar{\eta}(T-c_1d^2)},\frac{2\sqrt{\bar{\eta}}}{\sigma^A_{\min}}\right\} \cdot \left((c_1/2)d^2(tr(A)+d)+ d/2 \right) + c_1d^2\left(\frac{(1/2)tr(\Sigma)}{1+\sigma^A_{\min}}\right) 
    \end{split}.
 \end{align}

For the rank-deficient case, that is $rank(A)=r<d$,
\begin{align}
    \begin{split}
        \text{Regret}[m+1,T] \leq& \left(\frac{2d\sigma^A_{\max} \|\Sigma\|_F}{(\sigma^A_{\min}-\epsilon)^2}\right)\epsilon^2(T-m)^2+\sqrt{\epsilon} (T-m)\left(\frac{2\sqrt{\epsilon}\|\Sigma\|_F\sigma^A_{\max} \sqrt{2d}}{(\sigma^A_{\min}-\epsilon)(1+\sigma^A_{\min}-\epsilon)}+\frac{\sqrt{\epsilon}tr(\Sigma)}{(1+\sigma^A_{\min}-\epsilon)^2-1}\right) \\
        &+ \sqrt{\epsilon} (T-m)\left(\left[\frac{2d\epsilon^{3/2}}{(\sigma^A_{\min}-\epsilon)^2} + \frac{\sqrt{2d}\cdot \sqrt{\epsilon}}{\sigma^A_{\min}-\epsilon} \right]\|\Sigma\|_F + \max\left\{1,\frac{\sqrt{\epsilon}}{\sigma^A_{\min}}\right\}\cdot tr(\Sigma)\right). 
        \end{split}
\end{align}
Now, recall that $\epsilon = \frac{\|\bm{\eta}_{1:m}\|}{m\gamma^2} \leq \frac{\bar{\eta}}{\gamma^2}$ in our estimation+online optimization algorithm. The measurements rounds cost $m\gamma^2.$ We choose $\gamma^2 = \sqrt{\bar{\eta}}\max\left\{(T-m)^{2/3},\frac{2}{\sigma^A_{\min}}\right\}\implies \epsilon \leq \min\left\{\frac{1}{(T-m)^{2/3}},\frac{\sigma^A_{\min}}{2}\right\}$ and $m$ as the minimum number of rounds for consistency of the estimate $\hat{A}$, that is, $m = c_1\cdot rd$, where $r<d$ is the rank of $A$. This gives us a regret upper bound
\begin{align}
\begin{split}
    \regret_T \leq& \sqrt{\bar{\eta}}(T-c_1rd)^{2/3}\left(\frac{8d\sigma^A_{\max} \|\Sigma\|_F}{(\sigma^A_{\min})^2} +\frac{((4\sigma^A_{\max}+2)\sqrt{d} + 4\sqrt{2}d) \|\Sigma\|_F+(\sigma^A_{\min}+\sqrt{2})tr(\Sigma)}{(\sigma^A_{\min})^{1/2}} \right)\\
    &+ \sqrt{\bar{\eta}}\max\left\{(T-c_1rd)^{2/3},\frac{2}{\sigma^A_{\min}}\right\} \left((c_1/2)rd(tr(A)+d)+ d/2 \right) + c_1rd\left(\frac{(1/2)tr(\Sigma)}{1+\sigma^A_{\min}}\right)
\end{split}
\end{align}

\subsection{Proof of Theorem \ref{thm:lower-bound}}\label{proof:lower_bound}
Consider any algorithm of the form:
\begin{align}
    x_t = \hat{C}_t x_{t-1} + (I-\hat{C}_t)v_t
\end{align}
where $\frac{\epsilon}{2}\leq \sigma^{C_t-\hat{C}_t}_{\min} \leq  \sigma^{C_t-\hat{C}_t}_{\max} \leq \epsilon.$ The analysis of interpolation algorithms (Theorem 3.7) in \cite{BhuyanMukherjee24} can be extended to give the following naive regret bound (in the first step). Using the specific instance of $A = \bm{0}_{d\times d}$ (in the second step):
\begin{align}
    \regret_T &\geq \sum_{s=1}^T \frac{1}{2}\E\|x_{s-1}-v_s\|_{(\hat{C}_s-C_s)A(\hat{C}_s-C_s)+(\hat{C}_s-C_s)^2}^2\\
    &\geq \frac{\epsilon^2}{4} \sum_{s=1}^T \frac{1}{2}\E\|x_{s-1}-v_s\|_2^2\\
    &\geq \frac{\epsilon^2}{8} \sum_{s=1}^T \frac{1-(1-\epsilon)^{2s}}{\epsilon(2-\epsilon)}\\
    &\geq \frac{\epsilon T}{8} - \text{ constant}\\
\end{align}
Since exploration phase regret $\propto \frac{1}{\epsilon}$, the total regret is
\begin{align}
    \regret_T \geq \frac{1}{\epsilon} + \frac{\epsilon T}{8} - \text{ constant} = \Omega(\sqrt{T})
\end{align}

\section{The Fundamental Trade-off}\label{appnedix_sec:main_example}
\label{sec:fundamental}
    Consider the action space $\R^{d}$ where the player starts at the origin. The time horizon is fixed as $T=2d-1$ and underlying $A$ matrix is diagonal with $d$ distinct positive entries. Now, the environment supplies a sequence of minimizers $\{v_t\}_{t=1}^{2d-1}$ such that for the first $d$ rounds $(v_t)_2=(v_t)_3 = \ldots = (v_t)_{d} = 0$, and starting with round $(d+1)$, at $t=(d+i)^{th}$ round, $v_t$ is such that $v_{t} - x_{t-1}$ is parallel to $e_{i+1}$ for $i \in \{1,\ldots,d-1\}.$ Such a sequence of minimizers can occur adversarially or stochastically $\left(\text{for example, } v_{d+i} \sim \mathcal{N}(x_{t-1},e_{i+1}e_{i+1}^T) \right)$.
    
    \paragraph{Option 1: Tracking without estimation:}
    Most online algorithms in the SOCO literature \cite{GoelLinWierman19,GoelWierman19,zhang2021revisiting,BhuyanMukherjee24,bhuyan2024optimal,lin2022decentralized} have the general form:
    \begin{equation}\label{eqn1}
        x^{\alg}_t = \argmin_{x\in \R^d} f_t(x) + c(x,x_{t-1}) + g(x,v_t,x_{t-1}),
    \end{equation}
    which for quadratic cost functions, place $x_t$ on the line between $x_{t-1}$ and $v_t$. This means that any robust online algorithm $\alg{}$ dictates that the player take a sequence of actions $(x^{\alg}_1,\ldots,x^{\alg}_d)$ parallel to $e_1$ for the first $d$ rounds. During these rounds, the bandit feedback model collects information:
    \begin{align*}
        \bigg\{ (1/2) \cdot(x^{\alg}_k - v_k)^T A(&x^{\alg}_k - v_k) +\eta_k \bigg\}_{k=1}^{d} \\
        &= \left\{ c_k A_{1,1}+\eta_k\right\}_{k=1}^{d}.
    \end{align*}
    At round $(d+1)$, $x^{\alg}_{d+1}$ is supposed to be on the line between $x_{d}$ and $v_{d+1}$, which is parallel to $e_2$. Consequently, $x_{d+1}$ has \textbf{direct dependence} on $A_{2,2}$, the second diagonal entry of the unknown matrix~$A$. The player now resorts to the above rank-1 data collected so far, as that is the only information it has on matrix $A.$ 
    
    However, as it turns out, the first $d$ rounds only generated information about $A_{1,1,}$, forcing the player to make a \textbf{\textit{blind guess}} regarding $x_{d+1}.$ In fact it gets worse, as this will happen repeatedly, with round $d+i$ requiring the value of $A_{i+1,i+1}$ for computing a robust action but the player having knowledge of only $\{A_{1,1,},\ldots,A_{i,i}\}$. 
    
    \paragraph{Option 2: Full focus on estimation:}
    Alternatively, the player could have spent rounds $\{2,3\,\ldots,d\}$ probing the hitting cost along the rest of the directions, collecting information on the entire matrix $A$. However, this mechanism is very costly due to noise present in the bandit feedback. Consider the round $k$ (and its previous round $k-1$) where the player is probing the hitting cost $f_t(x) = \frac{1}{2}(x-v_k)^T A(x-v_k)$ using action $x_k = v_k + \gamma\cdot u_k$. The cost this measurement incurs is:
    \begin{align*}
        f_k(x_k) +c(x_k,x_{k-1}) > \gamma^2\left(\frac{\lambda^A_{\min}}{2} + \frac{\|x_k-x_{k-1}\|_2^2}{2}\right)
    \end{align*}
    where estimation requires $\gamma$ and $u_k$ to satisfy the following conditions to avoid \textit{data redundancy}
    \begin{align*}
        \gamma^2 \gg \frac{2\eta_k}{\lambda^A_{\min}} &\to \text{high signal-to-noise ratio (SNR)}\\
        u_k \not{\|} u_{k-1} &\to \text{avoid directional redundancy}.
    \end{align*}
    The data collection mechanism needs to satisfy the Restricted Uniform Boundedness (RUB) property \cite{CaiZhang15}, for recovery of the original matrix $A$:
    \begin{align}
        C_1 \leq \frac{\frac{1}{m}\sum_{k=1}^m \frac{\gamma^2}{2}u_k^T A u_k }{\|A\|_F} \leq C_2 \text{ } \forall \text{ } A \in \R^{d\times d} \text{ having } rank(A)=r
    \end{align}
    For this to hold with high probability for any symetric matrix $A$ with rank $r$, $\{u_k\}_k$ have to sampled independently and identically (IID) from a directionally uniform distribution, often a normal distribution. This leads to:
    \begin{align}
        \|x_k-x_{k-1}\|_2^2 &> \frac{\gamma^2}{2}\|u_k-u_{k-1}\|_2^2 - \|v_k-v_{k-1}\|_2^2
    \end{align}
    which under expectation
    \begin{align}
        \E\|x_k-x_{k-1}\|_2^2 &> \gamma^2 - tr(\Sigma)
    \end{align}
    It is therefore established that each round of measurement cost the player to the tune of $\Theta(\gamma^2)$, with $m$ rounds of measurements leading to a \textit{linear regret} of 
    \begin{align}
        \text{Regret}_{\text{Estimation}}[1,m] \simeq \Theta(\gamma^2)\cdot m
    \end{align}
    Due to ignorance towards high hitting and switching costs, there is a significant deviation from optimal trajectory \cite{GoelLinWierman19,BhuyanMukherjee24}. Consequently, we have the following trade-off:
    \begin{quote}
        \centering \small
        ``\textit{Collecting high-fidelity rank-$1$ data for matrix estimation incurs high online costs but potentially pays off in the long-run. Solely following a robust online algorithm might have initial benefit but leaves the player vulnerable should it require knowledge of the matrix $A.$}''
    \end{quote}
This illustrates that SOQO and data-driven learning are orthogonal tasks and are extremely difficult to combine.
\end{document}